\documentclass{article}

\usepackage{amsmath, amssymb, amsthm}
\usepackage{mathtools}
\usepackage{bbm}
\usepackage{dsfont}

\usepackage[usenames,dvipsnames]{xcolor}

\usepackage{enumitem}

\usepackage{graphicx}
\usepackage[labelfont=bf]{caption}
\usepackage[format=hang]{subcaption}

\usepackage{booktabs, array}
\usepackage{multirow}
\usepackage{subcaption}

\newsavebox\CBox

\newcommand{\tbf}[1]{\sbox\CBox{#1}\resizebox{\wd\CBox}{\ht\CBox}{\textbf{#1}}}
\newcommand{\tpm}[1]{\small{$\pm{\, #1}$}}

\usepackage{ifthen}
\newcommand{\tv}[3][n]{%
    \ifthenelse{\equal{#1}{n}}{%
        \ifthenelse{\equal{#3}{}}{#2}{#2 \tpm{#3}}%
    }{\ifthenelse{\equal{#1}{b}}{%
        \ifthenelse{\equal{#3}{}}{\tbf{#2}}{\tbf{#2} \tpm{#3}}%
    }{\ifthenelse{\equal{#1}{g}}{%
        \ifthenelse{\equal{#3}{}}{\color{gray}#2}{\color{gray}#2 \tpm{#3}}%
    }{%
        \textcolor{red}{ERROR}%
    }}%
}}

\usepackage{listings}
\usepackage{fancyvrb}
\fvset{fontsize=\small}

\usepackage{natbib}
\usepackage[colorlinks,linktoc=all,pagebackref=true]{hyperref}
\usepackage[hypcap=true]{caption}
\hypersetup{citecolor=NavyBlue}
\hypersetup{linkcolor=NavyBlue}
\hypersetup{urlcolor=NavyBlue}
\usepackage[nameinlink,capitalise]{cleveref}
\creflabelformat{equation}{#2\textup{#1}#3}  

\lstdefinestyle{mystyle}{
    commentstyle=\color{OliveGreen},
    numberstyle=\tiny\color{black!60},
    stringstyle=\color{BrickRed},
    basicstyle=\ttfamily\scriptsize,
    breakatwhitespace=false,
    breaklines=true,
    captionpos=b,
    keepspaces=true,
    numbers=none,
    numbersep=5pt,
    showspaces=false,
    showstringspaces=false,
    showtabs=false,
    tabsize=2
}
\usepackage[acronym,nowarn,section,nogroupskip,nonumberlist]{glossaries}
\glsdisablehyper{}

\newacronym{mlp}{\textsc{mlp}}{Multilayer perceptron}

\newcommand{\bg}{\mathbf{g}}

\newcommand{\bst}{\boldsymbol{t}}

\newcommand{\bsx}{\boldsymbol{x}}
\newcommand{\bsy}{\boldsymbol{y}}
\newcommand{\bsz}{\boldsymbol{z}}

\newcommand{\calD}{{\mathcal{D}}}

\newcommand{\calF}{{\mathcal{F}}}

\newcommand{\calL}{{\mathcal{L}}}

\newcommand{\calX}{{\mathcal{X}}}
\newcommand{\calY}{{\mathcal{Y}}}

\newcommand{\bbE}{\mathbb{E}}

\newcommand{\bbR}{\mathbb{R}}

\theoremstyle{plain}
\newtheorem{thm}{Theorem}[section]

\newtheorem{prop}[thm]{Proposition}

\theoremstyle{definition}
\newtheorem{defn}{Definition}[section]

\theoremstyle{remark}

\newcommand{\dee}{\mathrm{d}}

\DeclareMathOperator{\KL}{D_\mathrm{KL}}

\def\[#1\]{\begin{align}#1\end{align}}
\newcommand{\norm}[1]{\left\lVert{#1}\right\rVert}
\newcommand{\spm}[1]{\scriptstyle{\pm#1}}
\usepackage{microtype}
\usepackage{graphicx}
\usepackage{booktabs} 

\usepackage{hyperref}

\usepackage[accepted]{icml2024}

\usepackage{amsmath}
\usepackage{amssymb}
\usepackage{mathtools}
\usepackage{amsthm}

\theoremstyle{plain}

\theoremstyle{definition}

\theoremstyle{remark}

\usepackage[textsize=tiny]{todonotes}

\icmltitlerunning{Variational Partial Group Convolutions for Input-Aware Partial Equivariance}

\begin{document}

\twocolumn[
\icmltitle{Variational Partial Group Convolutions \texorpdfstring{\\for Input-Aware Partial Equivariance of Rotations and Color-Shifts}{}}



\icmlsetsymbol{equal}{*}

\begin{icmlauthorlist}
\icmlauthor{Hyunsu Kim}{gsai}
\icmlauthor{Yegon Kim}{gsai}
\icmlauthor{Hongseok Yang}{soc}
\icmlauthor{Juho Lee}{gsai,aitrics}
\end{icmlauthorlist}

\icmlaffiliation{gsai}{Kim Jaechul Graduate School of AI, KAIST, Daejeon, South Korea}
\icmlaffiliation{soc}{School of Computing, KAIST, Daejeon, South Korea}
\icmlaffiliation{aitrics}{AITRICS, Seoul, South Korea}

\icmlcorrespondingauthor{Hyunsu Kim}{kim.hyunsu@kaist.ac.kr}
\icmlcorrespondingauthor{Juho Lee}{juholee@kaist.ac.kr}
\icmlcorrespondingauthor{Hongseok Yang}{hongseok.yang@kaist.ac.kr}

\icmlkeywords{Machine Learning, ICML}

\vskip 0.3in
]



\printAffiliationsAndNotice{}  

\begin{abstract}
Group Equivariant CNNs (G-CNNs) have shown promising efficacy in various tasks, owing to their ability to capture hierarchical features in an equivariant manner. However, their equivariance is fixed to the symmetry of the whole group, limiting adaptability to diverse partial symmetries in real-world datasets, such as limited rotation symmetry of handwritten digit images and limited color-shift symmetry of flower images. Recent efforts address this limitation, one example being Partial G-CNN which restricts the output group space of convolution layers to break full equivariance. However, such an approach still fails to adjust equivariance levels across data. In this paper, we propose a novel approach, Variational Partial G-CNN (VP G-CNN), to capture varying levels of partial equivariance specific to each data instance. VP G-CNN redesigns the distribution of the output group elements to be conditioned on input data, leveraging variational inference to avoid overfitting. This enables the model to adjust its equivariance levels according to the needs of individual data points. Additionally, we address training instability inherent in discrete group equivariance models by redesigning the reparametrizable distribution. We demonstrate the effectiveness of VP G-CNN on both toy and real-world datasets, including MNIST67-180, CIFAR10, ColorMNIST, and Flowers102. Our results show robust performance, even in uncertainty metrics.
\end{abstract}

\begin{figure}[t]
     \centering
     \begin{subfigure}[b]{0.21\textwidth}
         \centering
         \includegraphics[width=\textwidth]{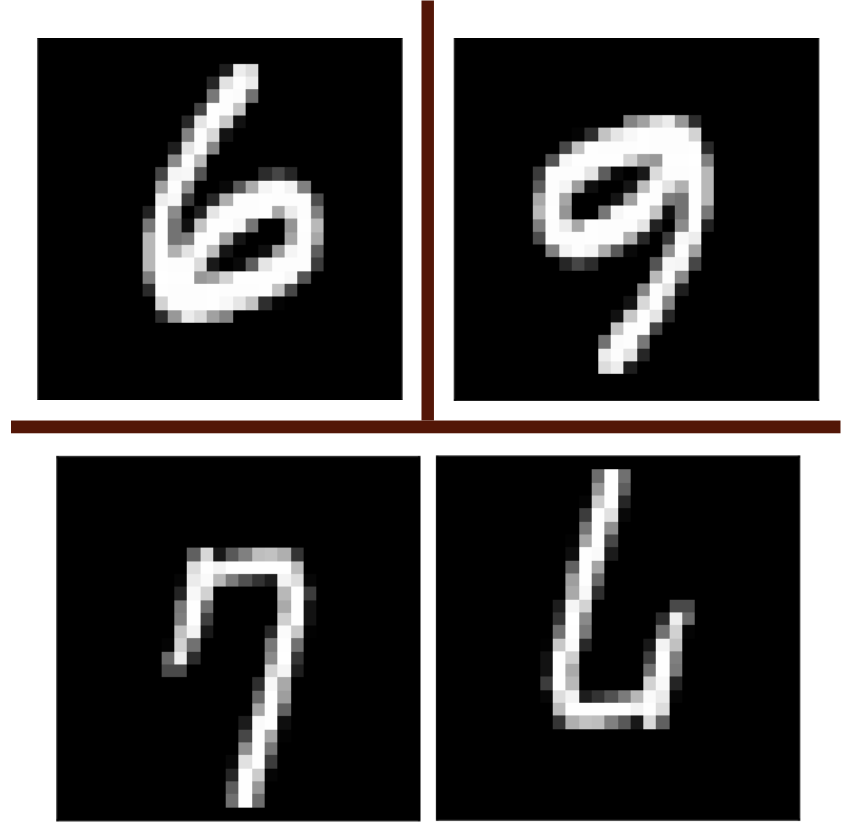}
         \caption{MNIST}
         \label{fig:partial1}
     \end{subfigure}
     \hfill
     \begin{subfigure}[b]{0.23\textwidth}
         \centering
         \includegraphics[width=\textwidth]{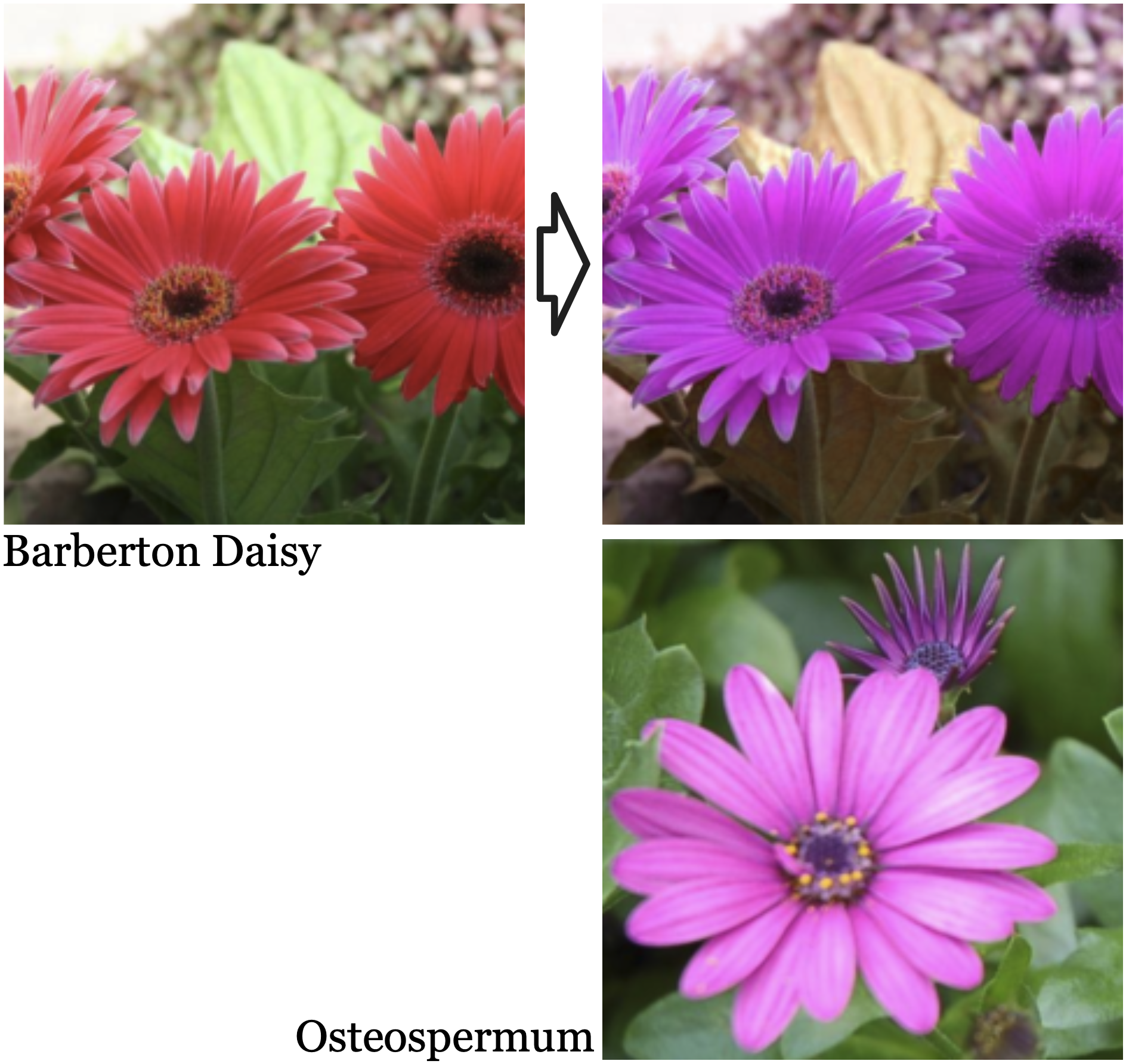}
         \caption{Flowers102}
         \label{fig:partial2}
     \end{subfigure}
        \caption{Illustrative example of partial equivariance. (a) $180^\circ$-rotation of 6 is regarded as 9 but 7 is not. (b) The color-shifted image of Barberton Daisy looks similar to Osteospermum.}
        \label{fig:partial}
\end{figure}
\section{Introduction}
\label{sec:main:introduction}

Convolutional Neural Networks (CNNs) have demonstrated remarkable success in numerous computer vision tasks, owing to their ability to capture hierarchical features in an equivariant manner. Other approaches, such as Group Equivariant CNNs (G-CNNs) \citep{cohen2016group, cohen2017steerable, weiler2019general, romero2022ckconv}, extend equivariance to various symmetry groups, enhancing model robustness across different transformations. However, a limitation arises from the rigidity of these models, as the choice of the equivariance group is fixed a priori.

In real-world scenarios, datasets often exhibit equivariance to diverse types of transformations, and the nature of equivariance might not be the same across all data instances. For example, in the classification of handwritten images like MNIST, images of 6 or 9 may be described more naturally by invariance to partial rotations between $-90^\circ$ and $90^\circ$, while a $180^\circ$ rotation might distort the classification between 6 and 9, as shown in \cref{fig:partial1}. In contrast, the other digits, ${0,1,2,3,4,5,7,8}$, may possess full equivariance to rotation. The challenge then lies in developing a neural network architecture that adapts the level of equivariance to the specific needs of the data.

Existing efforts have addressed this issue, such as Partial G-CNN \citep{romero2022learning}, which learns varying levels of equivariance at different layers, or Relaxed G-CNN \citep{wang2022approximately, ouderaa2022relaxing}, which incorporates relaxed kernel design. In particular, Partial G-CNN restricts the distribution of the output group space to break full equivariance. They introduce a convolution layer with a distribution whose support domain does not cover all group elements, effectively breaking equivariance. While this method has shown promising results, it imposes the same level of equivariance for all data points.

In this paper, we introduce a new group equivariant convolution that captures different levels of partial equivariance in a data-specific manner. We redesign the distribution of output group elements in Partial G-CNN to be conditioned on the input. For efficient computation, the data-dependent conditional distribution refers to features extracted from the previous layer, as these contain information about the input data. To train the conditional distribution without overfitting, we adopt Variational inference, treating the group elements in each layer as random variables. Thus, the problem becomes maximizing the evidence lower bound (ELBO), consisting of the log-likelihood for classification and the Kullback-Leibler (KL) divergence between the conditional distribution and a certain prior for regularization. Therefore, while the conditional distribution is regularized towards full equivariance, if full equivariance is harmful for the given data, it modifies the distribution to provide partial equivariance. Additionally, we address the unstable training issue in discrete group equivariance, which Partial G-CNN suffers from, by redesigning the reparametrizable distribution of the group elements. Our method, called Variational Partial G-CNN (VP G-CNN), shows promising results in terms of test accuracy and uncertainty metrics. It also demonstrates the ability to detect different levels of equivariance for each data point in one toy dataset, MNIST67-180, and three real-world datasets: CIFAR10, ColorMNIST, and Flowers102.

To sum up, our contributions can be summarized as follows:
\begin{enumerate}[itemsep=0mm]
    \item We propose input-aware partially equivariant group convolutions, which capture different levels of equivariance across data based on variational inference.
    \item We resolve the unstable training issue of discrete group equivariance involved in Partial G-CNN by redesigning the reparametrizable distribution for the discrete groups.
    \item We demonstrate promising results on real-world datasets: CIFAR10, ColoredMNIST, and Flowers102, alongside demonstrating strong calibration performance.
\end{enumerate}
\vspace{5em}
\section{Preliminaries}
\label{sec:main:background}

\subsection{Group Equivariance and Partial Equivariance}

A representation of a group $G$ on a Euclidean space $\mathbb{R}^n$ can be defined as a function $\rho$ mapping $G$ to the general linear group on $\mathbb{R}^n$ (i.e., the group of invertible $n\times n$ matrices with matrix multiplication as group composition and identity matrix as identity element), ensuring that $\rho$ preserves the composition operator and the identity element of the group. When we possess representations of a group $G$ in Euclidean spaces $\calX$ and $\calY$, denoted as $\rho_\calX$ and $\rho_\calY$ respectively, a function $\Phi : \calX \to \calY$ is termed \emph{equivariant to $G$} if, for all $g \in G$ and $\bsx \in \calX$, the following condition holds:
\[
\Phi\big(\rho_{\cal{X}}(g)(\bsx)\big) = \rho_{\cal{Y}}(g)\big(\Phi(\bsx)\big).
\]
In simpler terms, this condition implies that $\Phi$ does not actively utilize information that can be altered by group elements $g$.

As a more general concept, partial group equivariance, or \emph{partial equivariance}, can be defined as follows:


\begin{defn}[$(S, \varepsilon, G)$-Partial Equivariance]
\label{defn:partial-equivariance}
Let $\Psi: \calX \to \calY$ be a function and $G$ be a group acting on $\calX$. The function $\Psi$ is partially $G$-equivariant with respect to a subset $S \subseteq \calX$ and an error threshold $\varepsilon > 0$ if the following holds,
\[
&\sup_{g\in G}\norm{\Psi(\rho_\calX(g)(\bsx))-\rho_\calY(g)\big(\Psi(\bsx)\big)}=0, \quad \bsx\in S,\\
&\nonumber\sup_{g\in G}\norm{\Psi(\rho_\calX(g)(\bsx'))-\rho_\calY(g)\big(\Psi(\bsx')\big)}\leq\varepsilon, \quad \bsx'\in \calX \setminus S,
\] that is, it is equivariant on a given subset $S$ and approximately equivariant outside $S$.
\end{defn}

The set $S$ is determined with respect to the given dataset and group, typically defined as a subset of $\calX$ that excludes certain inputs known to possess specific symmetries. For example, in the MNIST dataset with respect to the $SO(2)$ group, subset $S$ includes digit images other than 6 and 9. Notice that for $\bsx \in S$, the function $\Psi$ must exhibit full equivariance, while for $\bsx \notin S$, it must exhibit $\varepsilon$-approximate equivariance. This definition ensures that equivariance is enforced on a specific subset $S$ of the domain, while allowing for $\varepsilon$-approximate equivariance with respect to inputs outside $S$.

\begin{defn}[$(C, \varepsilon, G)$-Partial Equivariance on Feature Map]
\label{defn:partial-equivariance-feature}
Let $G$ be a group acting on $\calF$ and $\Phi: \calF \to \calF$ be a map between functions $f:G\to\bbR^d$ representing input feature maps on group $G$. The function $\Phi$ is partially $G$-equivariant with respect to a subset $C \subseteq \calF$ and an error threshold $\varepsilon > 0$ if for all $u\in G$, it satisfies that:
\[
&\sup_{g\in G}\norm{\Phi(\calL_gf)(u)-(\calL_g\Phi(f))(u)}=0, \quad f\in C,\\
&\nonumber\sup_{g\in G}\norm{\Phi(\calL_gf')(u)-(\calL_g\Phi(f'))(u)}\leq\varepsilon, \quad f'\in \calF \setminus C
\]
where $\calL_g$ is the group representations of the group element $g$.
\end{defn}

\subsection{G-CNN and Partial G-CNN}

The convolutional layers of CNNs for an image can be described in terms of a function  $f:\bbR^2\to\bbR^3$ that maps the position of a pixel to its RGB vector and represents the input image, and a kernel $k:\bbR^2\to\bbR^{3 \times d}$, where $d$ is the output feature dimension. They output $(k*f) : \bbR^2 \to \bbR^d$ defined by $(k * f)(\bsy)=\int_{\bbR^2}k(\bsx-\bsy)f(\bsx)\dee\bsx$. The convolutional neural network exhibits translation equivariance due to the property $\calL_g(k*f)=k*\calL_g f$, where $\calL_g$ denotes a translation (shift) operation of image pixels: $\calL_g f(\bsx)=f(\bsx-\bst)$. Likewise, the convolutional layers of G-CNN utilize the equivariance property of the convolution operation on an extended space defined on a certain group $G$, which may include a translation group.

\paragraph{Lifting convolution.} We want to do the group convolution on a group $G$, but the input like an image is typically a map defined on a space $E \subseteq \bbR^m$ and so it needs to be lifted to a map from the group $G$. The lifting convolution performs this lifting. If there is an embedding of $E$ to $G$ so that $E$ can be regarded as a subgroup of $G$, we have, for an input feature map $f:E\to\bbR^3$ and a kernel map $k:G\to\bbR^{3\times d}$, the following lifting convolution $k *_\mathrm{lift} f : G \to \bbR^d$: for all $u\in G$,
\[
\label{eq:lifting-conv}
(k*_\mathrm{lift}f)(u) = \int_{v\in E}k(v^{-1}u)f(v)\dee\mu_E(v)
\] 
where elements in $E$ are viewed as group elements in $G$, and $\mu_E$ is the restriction of the left Haar measure of $G$ to the subgroup $E$.  Under an appropriate condition, the lifting convolution defined on group $G$ is equivariant to the group $G$, i.e. for all $g\in G$,
\[
(k*_\mathrm{lift}\calL_gf)(u)=\calL_g(k*_\mathrm{lift}f)(u),
\] where $\calL_gf(u) = f(g^{-1}u)$.

\paragraph{Group convolution.} The group convolution generalizes the regular convolution for equivariances with respect to general groups. Once the inputs are feature maps from $G$, the group equivariant convolution for an input feature map $f: G\to\bbR^d$ and a kernel $k: G\to\bbR^{d\times n}$, where $n$ is the output feature dimension, is defined as follows:
\[
\label{eq:group-conv}
(k*f)(u) = \int_{v\in G}k(v^{-1}u)f(v)\dee\mu_G(v),
\] where $\mu_G$ is the left Haar measure of the group $G$. Similarly to the regular convolution, the group convolution is $G$-equivariant, that is, $k*\calL_gf=\calL_g(k*f)$.

\paragraph{Partial group convolution.} Inspired by Augerino \citep{benton2020learning}, Partial G-CNN \citep{romero2022learning} introduced a partially equivariant group convolution whose output feature space is determined by a distribution $q(u)$, where $u\in G$. It modified the group convolution as follows:
\[
(k*f)(u) = \int_{v\in G}q(u) k(v^{-1}u)f(v)\dee\mu_G(v).
\] 
For instance, when $G$ is the 2-dimensional rotation group $SO(2)$ with radian values in $[-\pi,\pi]$, the distribution $q(u)$ can be defined as the push forward of the exponential map $\exp:\bg\to G$ of the distribution $\mathrm{Unif}[R(-\theta),R(\theta)]$ on the Lie algebra $\bg$, where $\theta$ is a learnable parameter on radian space and $R:\bbR\to so(2)$, and represents the maximum possible rotations in $\bbR^2$. That is,
\[
u = \exp(t),\quad t\sim\mathrm{Unif}[R(-\theta),R(\theta)].
\]
If the full equivariance (i.e. $\theta=\pi$) is harmful for training, the model modifies the $\theta$ to be less than $\pi$. However, Partial G-CNN fails to guarantee the partial equivariance for a non-empty $S$ in~\cref{defn:partial-equivariance}, when $\theta < \pi$. This is because, when $\theta$ becomes less than $\pi$, Partial G-CNN loses equivariance to $G$ for \emph{all} $\bsx\in \calX$. This departure from equivariance violates the condition specified for a subset S if $S \neq \emptyset$. The model either exhibits full equivariance when $\theta=\pi$ or broken equivariance when $\theta<\pi$. For convenience, we omit the exponential map and mapping $R$ when we describe the distribution of group elements, and write $q(u;\theta) = \mathrm{Unif}[-\theta,\theta]$ or $q(u)=\mathrm{Unif}[-\theta,\theta]$.

\paragraph{Color equivariance $H_m$.} We aim to achieve equivariance not only with respect to the standard group SE(2), but also concerning color shifts. In \citep{lengyel2023color}, color equivariance is defined as being equivariant to changes in hue. It is explained that the Hue-Saturation-Value (HSV) color space represents hue using an angular scalar value, and shifting hue involves a straightforward additive adjustment followed by a modulo operation. When translating the HSV representation into the three-dimensional RGB space, a hue shift corresponds to a rotation along the $(1,1,1)$ diagonal vector. Color equivariance is established in terms of a group by defining $H_m$, which consists of multiples of $360/m^\circ$ rotations around the $(1,1,1)$ vector in $\bbR^3$. $H_m$ is a subgroup of $SO(3)$, the group of all rotations about the origin in $\bbR^3$. The group operation is matrix multiplication, acting on the continuous space of RGB pixel values in $\bbR^3$. Consequently, color-equivariant convolutions can be constructed using discrete $SO(3)$ convolutions when the RGB pixels of an image are treated as $\bbR^3$ vectors forming three-dimensional point clouds.
\section{Variational Partial G-CNN}
\label{sec:main:method}

\subsection{Input-Aware Partial Convolution}

In order to achieve partial equivariance defined in~\cref{defn:partial-equivariance},
we need to make the distribution $q(u)$ input-aware, and design $q(u|\bsx)$ for each input $\bsx$. One approach is to put $q(u|\bsx)$ for every layer, but doing so would be memory-inefficient, especially for the continuous group convolutions. This approach requires retaining the group elements sampled from $q(u|\bsx)$ for all convolution layers during feed-forwarding.

Therefore, for partial equivariance, our new convolution at layer $l+1$ uses $q(u|f^{(l)})$ where $f^{(l)}$ is the output of the previous layer $l$. Since as a feature, $f^{(l)}$ contains information about the input data, this scheme has a potential to identify data-specific equivariance, while being memory-efficient. Concretely, we modify the convolutions in~\cref{eq:lifting-conv,eq:group-conv}  as follows:
\[
\label{eq:bayes-partial}
\begin{aligned}
(k*_\mathrm{lift}f)(u) &= \int_{v\in E}q(u|f)k(v^{-1}u)f(v)\dee\mu_E(v),\\
(k*f)(u) &= \int_{v\in G}q(u|f)k(v^{-1}u)f(v)\dee\mu_G(v).
\end{aligned}
\]
The distribution $q(u|f)$ here must be partially equivariant in order to achieve partial equivariance in these convolutions. For example, if the input $f$ is the image of digit 7 or 8, which require full equivariance to $SO(2)$, $q(u|f)$ can be just the uniform distribution for all rotations in $\bbR^2$: $q(u|f)=\mathrm{Unif}[-\pi,\pi]$. Note that in this case, $q(u|f)$ is equivariant to $SO(2)$ in the following sense: $q(u|f) =q(gu|\calL_gf)$ for all $g\in SO(2)$. On the other hand, for the images of digit 6 or 9, which require only partial equivariance to $SO(2)$, $q(u|f)$ can be a uniform distribution with a narrower range, such as $\mathrm{Unif}[-\pi/2,\pi/2]$, or just a dirac-delta distribution $\delta(u)$. Note that in this case, $q(u|f)$ may fail to satisfy the equivariance condition, i.e., $q(u|f)\neq q(gu|\calL_gf)$ for some $g \in G$. 
The next proposition gives one sufficient condition for ensuring partial equivariance of our convolutions:

\begin{prop}
\label{prop:partial-dist}
    Assume that the conditional distribution $q(u|f)$ is partially equivariant with respect to a group $G$ and an equivariant subset $C\subseteq\calF$ in the following sense: 
    \[
    &\sup_{g\in G}\norm{q(u|f)-q(gu|\calL_gf)}=0,\quad f\in C,\nonumber\\
    &\sup_{g\in G}\norm{q(u|f')-q(gu|\calL_gf')}\leq\varepsilon,\quad f'\in \calF\setminus C,
    \] where $\calL_gf(u) = f(g^{-1}u)$, and kernel $k$ and input $f$ of the group convolutions defined in~\cref{eq:bayes-partial} are bounded. Then, the group convolutions are also partially equivariant to $G$ and $C$.
\end{prop}

The proof is presented in \cref{proof:partial-dist}. For continuous groups, the integrals in the convolutions are intractable, so we typically employ Monte Carlo approximation to estimate the convolution operation by uniformly sampling from the Haar measure $\dee\mu_G$. Thus, the approximate partially equivariant group convolution is determined as follows:
\[
(k*f)(u_j) = \sum_{v_i}q(u_j|f)k(v_i^{-1}u_j)f(v_i).
\]
Now, we describe how the distribution $q(u|f)$ can be trained and implemented using variational inference with the reparametrization trick.

\subsection{Variational Inference of $q(u|f)$}
\label{sec:partial-distribution}

\begin{figure}[t]
    \centering
    \includegraphics[width=0.45\textwidth]{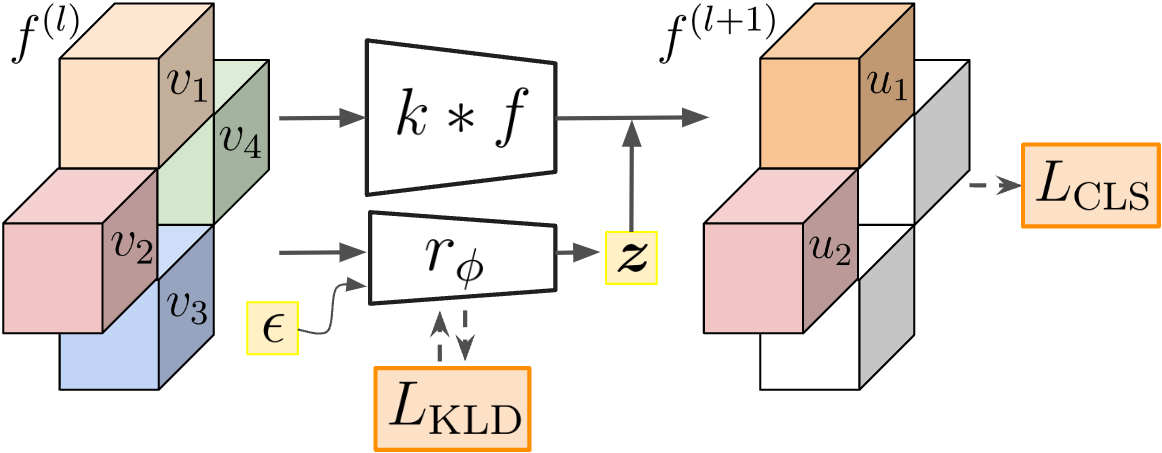}
    \vspace{-3mm}
    \caption{Architecture of Variational Partial Group Convolutions. The colored boxes are the features at each layer and the white boxes are zero features removed out by the distribution $q(u|f)$, where $u=r_\phi(f,\epsilon)$.}
    \label{fig:architecture}
\end{figure}

If we train $q(u|f)$ with only the classification loss, since it encompasses all features $f$, it may overfit by tending to become another classifier itself, leading to a trivial distribution. To prevent this situation, we adopt variational framework to train the distribution $q(u|f)$. Our goal is to maximize the log-likelihood $\log p(y|\bsx)$ for $\bsx,y$ from a dataset $\calD$ and it can be described as follows:
\[
\label{log-likelihood}
\lefteqn{\log p(y|\bsx)}\\
=& \int_{G}\log p(y|f^{(0)}, u^{(1)},\ldots,u^{(L)})\Pi_{l=1}^{L}p(u^{(l)})\dee\mu_G(u^{(l)})\nonumber,
\]
where $\bsx = f^{(0)}$, $L$ is the number of layers of the model, and $u^{(l)}$ is the output group elements at layer $l$.

To estimate the approximate posterior $q(u^{(l)}|f^{(l)})$ at layer $l$, we maximize the evidence lower bound (ELBO) of the log-likelihood in~\cref{log-likelihood}:
\[
\lefteqn{L_\mathrm{VP} =}\nonumber\\
&\bbE_{\{u^{(l)}\}_{l=1}^L}\bigg[\log\frac{p(y|f^{(0)},\{u^{(l)}\}_{l=1}^L)\Pi_{l=1}^Lp(u^{(l)})}{\Pi_{l=1}^L q(u^{(l)}|f^{(l)})}\bigg],
\] where the expectation is over $\{u^{(l)}\}_{l=1}^L\sim \Pi_{l=1}^L q(u^{(l)}|f^{(l)})$. Then, $\bbE_\calD[\log p(y|\bsx)]\geq L_\mathrm{VP}$ and by maximizing $L_\mathrm{VP}$, we can maximize the log-likelihood indirectly. The approximate posterior $q(u^{(l)}|f^{(l)})$ is the partially equivariant distribution shown in~\cref{eq:bayes-partial}. 

\begin{figure}[t]
    \centering
    \includegraphics[width=0.36\textwidth]{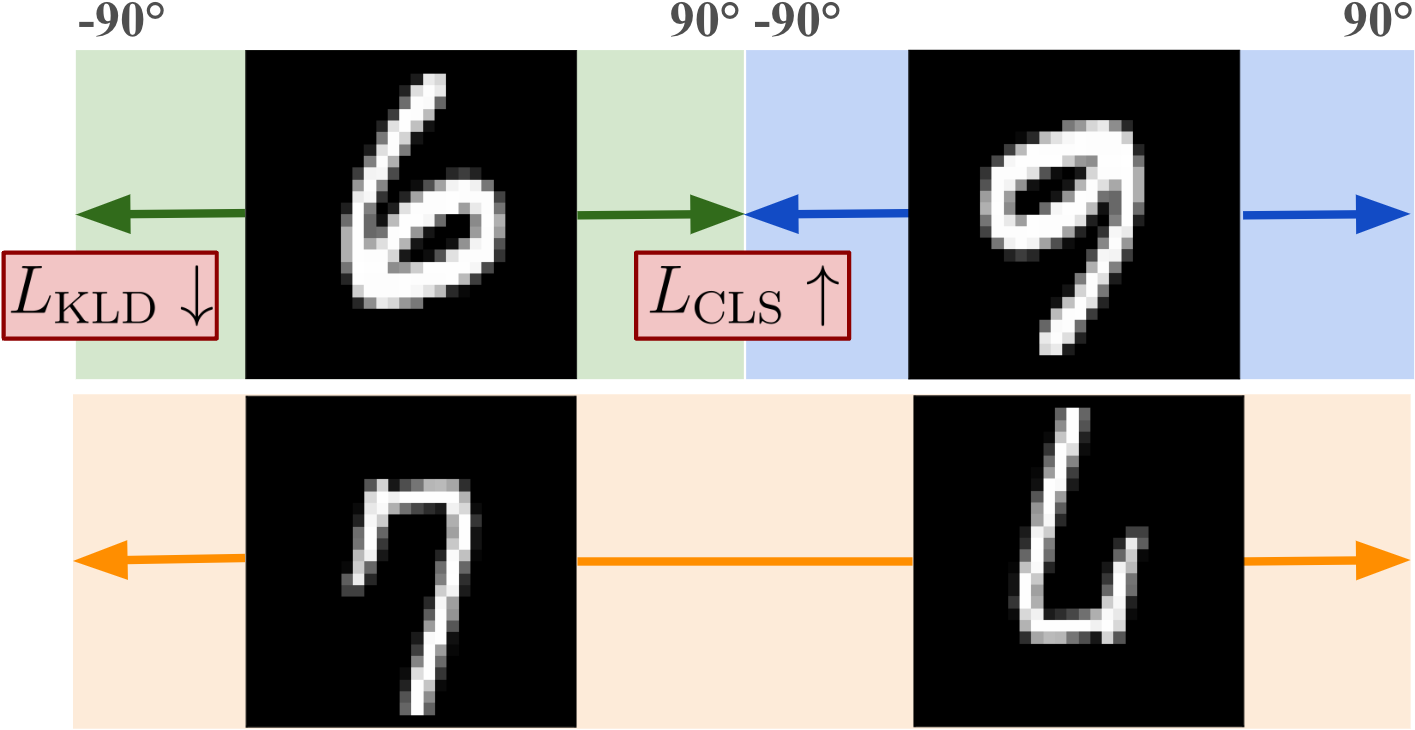}
    \vspace{-3mm}
    \caption{As the $L_\mathrm{KLD}$ increases, the distribution $p(u|f)$ expands, but upon reaching a certain point where $L_\mathrm{CLS}$ is affected, the distribution becomes constrained.}
    \label{fig:principle}
\end{figure}

In fact, ELBO can be viewed as two components consisting of maximizing likelihood for classification and minimizing Kullback-Leibler (KL) divergence between the approximate posterior $q(u^{(l)}|f^{(l)})$ and prior $p(u^{(l)})$ for regularization.
\[
\lefteqn{L_\mathrm{VP} = L_\mathrm{CLS} - \sum_{l=1}^L L_\mathrm{KLD}^{(l)},}\nonumber\\
&L_\mathrm{CLS} = \bbE_{u^{(1)},\ldots,u^{(L)}}\big[\log p(y|f^{(0)},u^{(1)},\ldots,u^{(L)})\big]\nonumber\\
&L_\mathrm{KLD}^{(l)} = \KL\Big(q(u^{(l)}|f^{(l)}) \big\Vert p(u^{(l)})\Big).
\]
The prior distribution is set to be a uniform distribution in which the probabilities of every group elements are the same, which corresponds to the full equivariance. Therefore, the model regularize $q(u|f)$ to preserve the full equivariance but if the full equivariance is harmful for training, it adjust the distribution $q(u|f)$ far from the uniform distribution. This principle is illustrated in~\cref{fig:principle}. In practice, a hard regularization of the KL divergence is possible to disturb training of the target model. Therefore, we control strength of $L_\mathrm{KLD}$ by adopting a coefficient $\lambda\in[0,1]$,
\[
L_\mathrm{VP} = L_\mathrm{CLS} - \lambda \sum_{l=1}^L L_\mathrm{KLD}^{(l)}.
\] $\lambda$ is a hyperparameter that user can assign.

To efficiently train the distribution $q(u|f)$, we need to estimate the gradient of the loss with low variance. Thanks to reparametrization trick~\citep{kingma2014auto}, if we design the distribution possible to allow the backpropagation, we get estimates of the gradient with low variance. The gradient of ELBO can be estimated as follows:
\[
\label{eq:reparam}
\nabla_{\theta} L_\mathrm{CLS} &= \bbE_{\epsilon^{(1)},\ldots,\epsilon^{(L)}}\big[\nabla_{\theta} \log p_\theta(y|\bsx,\bsz^{(1)},\ldots,\bsz^{(L)})\big],\nonumber\\
\nabla_\phi L_\mathrm{KLD}^{(l)} &= \nabla_\phi \KL\Big(q_\phi(\bsz^{(l)}|f^{(l)}) \big\Vert p(\bsz^{(l)})\Big)\nonumber\\
&= \bbE_{\epsilon^{(l)}}\bigg[\nabla_\phi\log\frac{p(\bsz^{(l)})}{q_\phi(\bsz^{(l)}|f^{(l)})}\bigg],
\] where $\bsz^{(l)} = r_\phi^{(l)}(f^{(l)}, \epsilon^{(l)})$ and $\theta,\phi$ are the parameters of the classifier $p_\theta$ and the group element encoder $r_\phi$, respectively, and $\theta$ includes $\phi$ because the classifier shares parameter with the encoder. The architecture of the input-aware partial group convolution is summarized in~\cref{fig:architecture}.

The partially equivariant distribution $q(u^{(l)}|f^{(l)})$ is sampled by uniformly drawing noise $\epsilon$ and feed-forward through the group element encoder $r_\phi$. The reparametrizable encoder is designed differently across the continuous group and the discrete group. Although our method is able to apply multi-dimensional continuous and dicrete groups when appropriate distribution is defined, we narrow down the scope to the continuous two-dimensional rotation group $SO(2)$ and the discrete color-shift group $H_m$, which are widely tackled in the examples of the partial equivariance.

\paragraph{Rotation $SO(2)$ (continuous).} Similar to Partial G-CNN~\citep{romero2022learning}, we can define $q(u|f)$ a uniform distribution $\mathrm{Unif}[-\theta, \theta]$ but $\theta$ is calculated from encoding of the input feature, $\theta=e_\phi(f),\,\theta\in [0, 1]$, then $r_\phi(f,\epsilon)$ is described as
\[
\label{eq:so2}
r_\phi(f,\epsilon)= \epsilon\pi \cdot e_\phi(f),\quad\epsilon\sim\mathrm{Unif}[-1,1].
\] If $\theta=1$, the probabilities of all group elements are the same, while if $\theta=0$, the distribution becomes a dirac-delta distribution whose value is non-zero only at zero-rotation. This distribution is reparametrizable so we can estimate the gradient as in~\cref{eq:reparam} with low variance.

\paragraph{Color-shift $H_m$ (discrete).} The color-shift group $H_m$ has $m$ number of group elements and each represents $360/m^\circ$ rotations around the $(1,1,1)$ vector in the three-dimensionl RGB vector space. To sample group elements in such a discrete group, Partial G-CNN utilizes Gumbel-Softmax trick~\citep{maddison2017concrete} with Straight-Through estimation but it suffers from unstable training~\citep{romero2022learning}. We observe that the distribution $p(u)$ with learnable parameters irregularly change their distribution during training and this may be due to the multi-modality of Gumbel-Softmax. Therefore, we propose another probability distribution that samples the discrete group without Gumbel-Softmax and mimick the distribution described in the continuous group.

For sampling, we first encode the input feature to $\theta=e_\phi(f),\,\theta\in[0,\infty)$ and sample $\{\epsilon_i\}_{i=1}^m$ from a discrete uniform distribution $\mathrm{Unif}\{1,2,\ldots,m\}$, corresponding to the uniform distribution in the continuous group. Then, we compute importance weights for each $\epsilon_i$ as
\[
\label{eq:importance}
w_i = \frac{\exp(\epsilon_i/\theta)}{\sum_{i=1}^m\exp(\epsilon_i/\theta)}.
\] Here, $\theta$ determines smoothness of the softmax function across each $i$th component; if $\theta$ is large enough, $w_i$ converges to almost uniform. Now using Straight-Through estimator, we select which group element in $\{u_i\}_{i=1}^m$ should be non-zero.
\[
\label{eq:ste}
q(u_i|f) =
    \begin{cases}
      1, & \text{if $w_i > \frac{1}{m}-\eta$,}\\
      0, & \text{otherwise,}\\
    \end{cases} 
\] where $\eta\in[0,1/m]$ is a hyperparameter that determines how easy to be selected as non-zero. As $\theta$ increases, the difference in magnitude between $w_i$ decreases, and more elements surpass the threshold. Conversely, as $\theta$ decreases, the difference in magnitude between $w_i$ increases, and fewer elements surpass the threshold. This principle is analogous to the distribution of continuous groups. For example, if $\eta$ is zero, $w_i$  should be greater than $1/m$ to be non-zero so it always select only one group elements, whereas if $\eta$ is $1/m$, it always select every elements in the group. The model trains value of $\theta$ so that it decides how many group elements are appropriate for given input. For instance, $m=3$, $\eta=7/12$, $\{\epsilon_i\}_{i=1}^m = \{3,2,1\}$, and then the threshold $1/m-\eta=0.25$. For $\theta=1$, $\{w_i\} = \{0.67, 0.24,0.09\}$ and $0.67$ is the only value larger than $0.25$, thereby only $u_1$ is selected. For $\theta=3$, $\{w_i\}=\{0.45,0.32,0.23\}$ and $0.45,0.32$ are above the threshold, thus $u_1$ and $u_2$ are selected. Since at least one of the softmax result in~\cref{eq:importance} for $m$ candidates should be greater than $1/m$, \cref{eq:ste} always selects at least one group elements.

\subsection{Implementation}

Utilizing the input-aware partial group convolution for every layers would be the best strategy to gain performance. However, there are limitations to performance improvement compared to the increase in parameters. Hence, throughout the experiments we set a portion of layers to be the input-aware partial convolution in a network. In fact, once at least one of the convolutional layers exhibits input-aware partial equivariance, the entire network becomes partially equivariant.
\begin{prop}
\label{prop:one-layer}
If at least one of the convolutional layers in a G-CNN is partially equivariant to a group $G$ and an equivariant subset $C\subseteq\calF$, and its activation functions are equivariant with respect to $G$ and $L$-Lipschitz continuous, and its kernel functions are bounded, then the entire G-CNN is also partially equivariant to $G$ and $C$.
\end{prop}
Its proof is described in~\cref{proof:one-layer}. For example, in the CIFAR10 dataset, we apply the input-aware partial group convolution in the lifting convolution and the last group convolution only. In the Flower102 dataset, we apply it in the last two group convolution only. In addition, we use light-weighted encoder $e_\phi$, which calculate $\theta$ as in~\cref{eq:so2,eq:importance}, consisting of two global average pooling layers, two one-dimensional convolution, and one linear layer. The detailed architecture is described in~\cref{sec:app:encoder}.
\section{Related Work}
\label{sec:main:relatedwork}

\textbf{Group equivariant networks.} G-CNN \citep{cohen2016group} proposed a convolutional neural network architecture ensuring equivariance to a group of input transformations, including translation, rotation, and reflection, thereby enhancing the model's ability to learn and generalize from data with inherent symmetries in a given dataset. Steerable CNN \citep{cohen2017steerable} introduced a framework for constructing rotation-equivariant convolutional neural networks, enabling efficient and flexible modeling of rotational symmetries in image data by leveraging the theory of group representations. $E(2)$-CNN \citep{weiler2019general} demonstrated constraints based on group representations, simplifying them to irreducible representations and providing a general solution for $E(2)$, thereby covering continuous group equivariance for images. CEConv \citep{lengyel2023color} extended equivariance from geometric to photometric transformations by incorporating parameter sharing over hue shifts, interpreted as a rotation of RGB vectors, offering enhanced robustness to color changes in images.

\textbf{Approximate equivariance.} RPP \citep{finzi2021residual} involved placing one equivariant neural network (NN) and one non-equivariant NN in parallel, with a prior imposed on the parameters of each NN. In contrast, PER \citep{kim2023regularizing} replaced the two components with a single non-equivariant NN and introduced a regularizer to drive the non-equivariant NN towards equivariance. Relaxed G-CNN \citep{wang2022approximately} introduced a small linear kernel to G-CNN, which slightly breaks the group equivariance of the model. In Partial G-CNN \citep{romero2022learning}, a distribution of group elements in the output was adopted, allowing group convolutions to consider only a subset of group elements in the hidden space.

\textbf{Input-aware automatic data augmentation.} MetaAugment \citep{zhou2021metaaugment} presents an efficient approach to learning a sample-aware data augmentation policy for image recognition by formulating it as a sample reweighting problem, where an augmentation policy network adjusts the loss of augmented images based on individual sample variations. AdaAug \citep{cheung2022adaaug} learns adaptive data augmentation policies in a class-dependent and potentially instance-dependent manner, addressing the limitations of methods like AutoAugment \citep{cubuk2019autoaugment} and Population-based Augmentation \citep{ho2019population} by efficiently adapting augmentation policies to specific datasets. InstaAug \citep{miao2023learning} learns input-specific augmentations automatically by introducing a learnable invariance module that maps inputs to tailored transformation parameters, facilitating the capture of local invariances. \citet{singhal2023learning} designed a method to capture multi-modal partial invariance by parameterizing the distribution of instance-specific augmentation using normalizing flows.
\section{Experiments}
\label{sec:main:experiment}

\begin{figure}[t]
     \centering
     \begin{subfigure}[b]{0.23\textwidth}
         \centering
         \includegraphics[width=\textwidth]{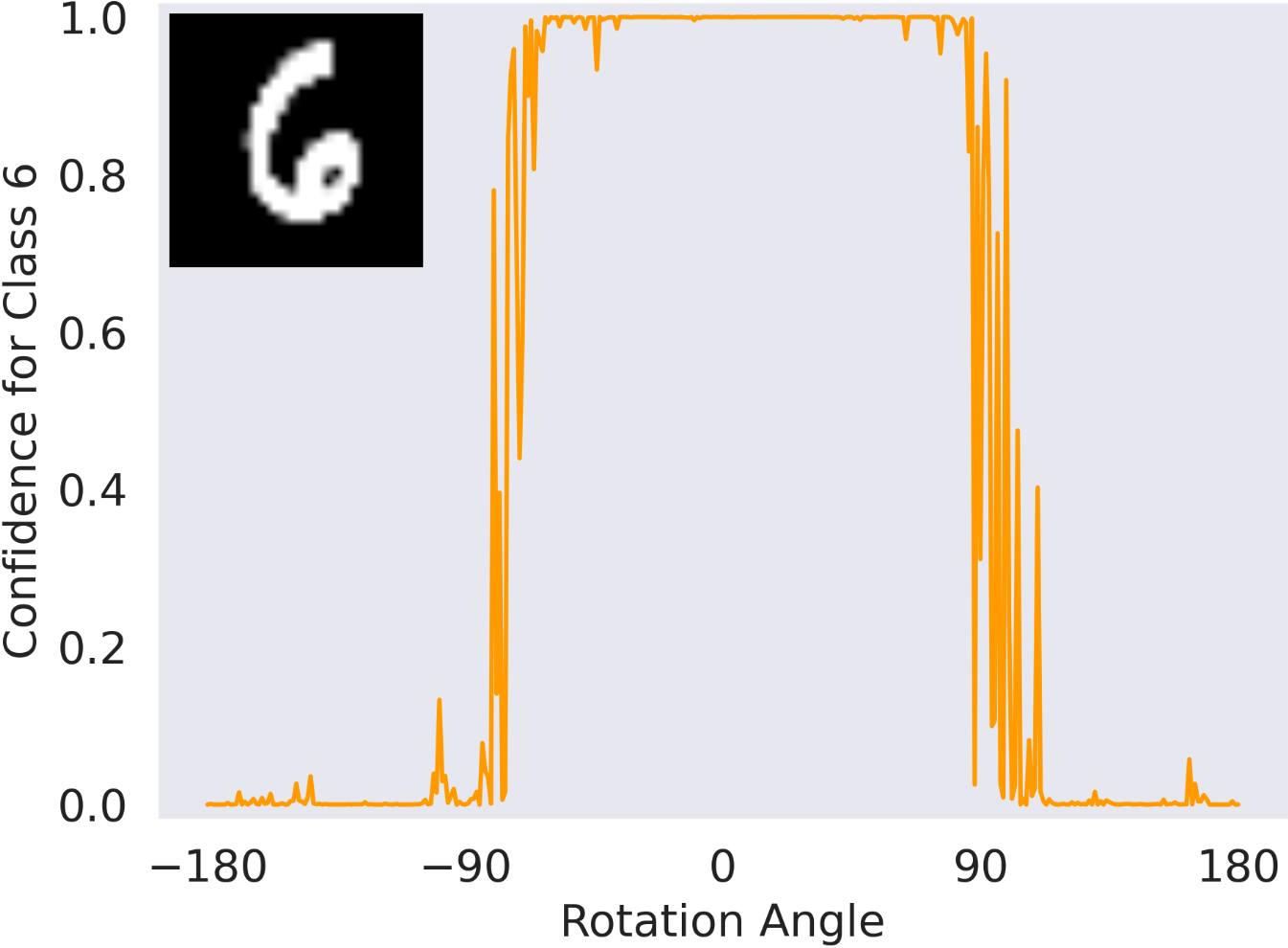}
     \end{subfigure}
     \hfill
     \begin{subfigure}[b]{0.23\textwidth}
         \centering
         \includegraphics[width=\textwidth]{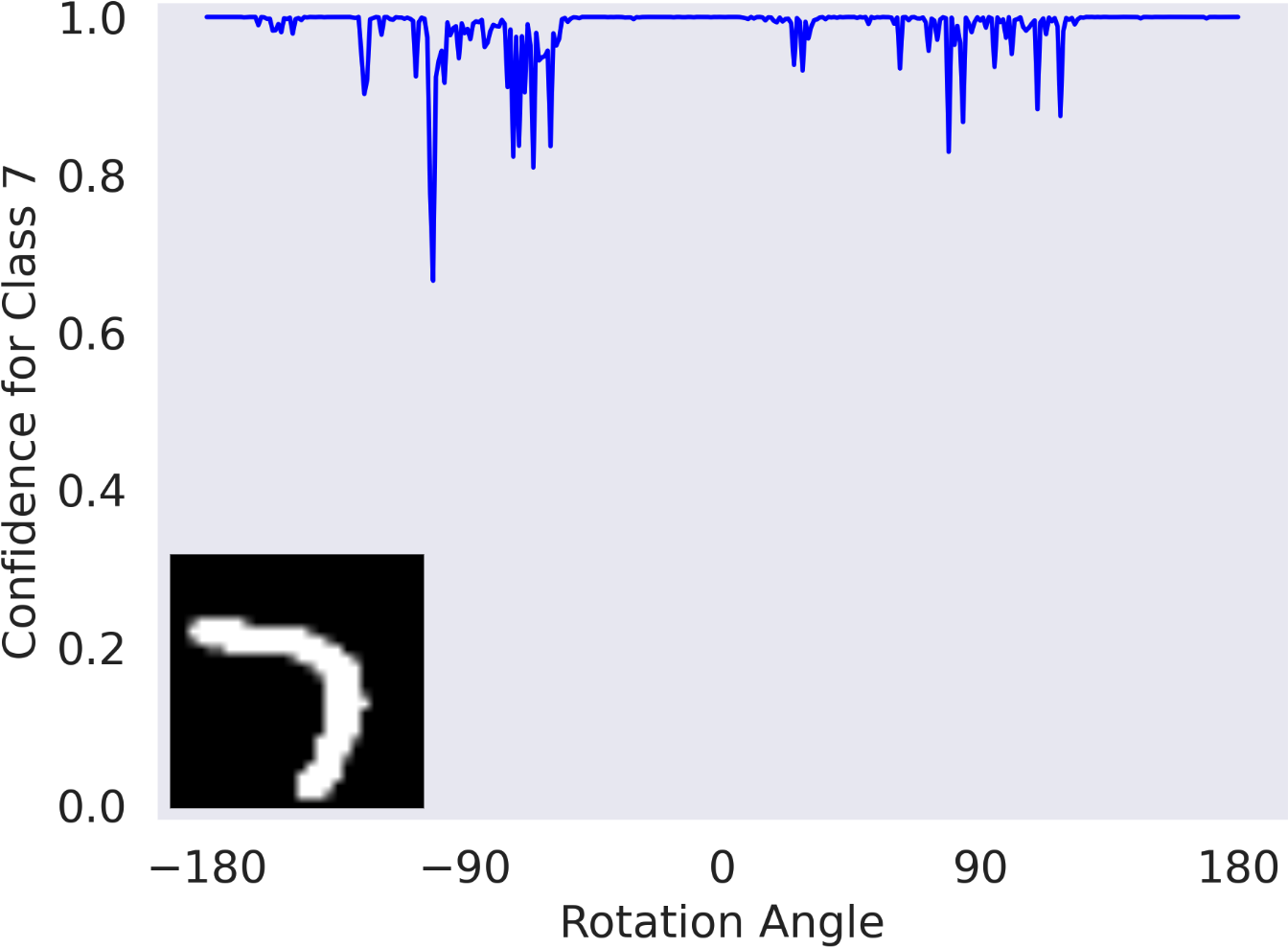}
     \end{subfigure}
     \vspace{-3mm}
        \caption{Partial equivariance trained on MNIST67-180. The x-axis represents the rotation angle of the input and the y-axis represents the model's confidence for the corresponding class. The model exhibits equivariance to rotations on semi-circle for image 6, whereas it shows full equivariance for image 7.}
        \label{fig:mnist67-180}
\end{figure}

In commonly addressed tasks, approximate equivariance often manifests in forms such as rotation and color shifts. To evaluate VP G-CNN's partial equivariance in rotations, we conduct experiments on two datasets: MNIST67-180 and CIFAR10 \citep{krizhevsky2009learning}. For color shifts, we assess performance on long-tailed colorMNIST, which exhibits full equivariance for color shifts but has imbalanced classes, and on Oxford Flower102 \citep{nilsback2008automated}, where partial equivariance is data-specific, as depicted in \cref{fig:partial2}. We compare our model with four baseline methods: ResNet (T(2)-CNN), G-CNN, Partial G-CNN, and InstaAug. InstaAug \citep{miao2023learning} is an AutoAugment technique that learns the appropriate distribution of augmentations for each data instance. Detailed hyperparameters used to train VP G-CNN and the baselines are listed in~\cref{sec:app:hyperparameters}. The source code demonstrating the experiements in colorMNIST and Flowers102 is available at \url{https://github.com/yegonkim/partial_equiv}.

\paragraph{Model architecture for $SE(2)$.} The group $SE(2)$ consists of translations $T(2)$ and rotations $SO(2)$. Similar to Partial G-CNN for $SE(2)$, we employ the extended version of G-CNN proposed by \citet{finzi2020generalizing}, Continuous Kernel Convolution (CKConv) \citep{romero2022ckconv}. However, we use the input-aware partial group convolution as defined in \cref{eq:bayes-partial}, and we parametrized the convolutional kernels $k$ as SIRENs~\citep{sitzmann2020implicit}. The overall structure is based on ResNet \citep{he2016residual} and it consists of one lifting convolution, two residual blocks, and one last linear layer. According to~\cref{prop:one-layer}, we apply the input-aware convolution on the lifting convolution and the last group convolution and the other convolutions are all partial group convolution of Partial G-CNN. We define $q(u|f)$ the straight-through distribution as proposed in~\cref{eq:so2}.

\paragraph{MNIST67-180 (toy dataset).} Inspired by MNIST6-180, as introduced in \citep{romero2022learning}, we created a new classification dataset named MNIST67-180. This dataset is derived from the MNIST handwritten dataset \citep{lecun2010mnist} and consists of images labeled as either 6 or 7, along with their corresponding $180^\circ$-rotated versions labeled as 9 and 7, respectively. Consequently, images of 6 should be classified as 6 within a rotation range of $[-90^\circ, 90^\circ]$, and as 9 within other angles of rotation. Meanwhile, images of 7 should always be classified as 7, regardless of the angle of rotation. We demonstrate the learned partial equivariance for some of the data.

We plot the probabilities of assigning the label 6 for image 6 and the label 7 for image 7 with respect to the test samples of MNIST67-180 rotated at whole angles in $[0^\circ,360^\circ]$. As shown in \cref{fig:mnist67-180}, the model learns to predict image 6 as 6 within the rotation range of $[-90^\circ,90^\circ]$, while it learns to predict image 7 as 7 within the rotation range of $[-180^\circ,180^\circ]$. This proves that our VP G-CNN learns an appropriate level of equivariance that varies for each type of data.

\begin{table}[t]
\small
\caption{Test accuracy on CIFAR10 with $SE(2)$-CNNs. \textit{P} and \textit{VP} denote that their architecture includes Partial and VP convolutional layers, respectively. \checkmark in the InstaAug column means the training is conducted with the augmentation of InstaAug.}
 \label{tab:test-rotation}
    \centering
 \resizebox{.4\textwidth}{!}{
    \begin{tabular}{ccccc}\toprule
        Group & \#Elems. & Partial & InstaAug  & CIFAR10 \\
        \midrule
        \multirow{2}{*}{$T(2) $} & \multirow{2}{*}{1}& \multirow{2}{*}{-}     & -       & $82.0 \spm 0.2$    \\
                               &                     &                        & \checkmark    & $81.9 \spm 0.4$    \\
        \midrule
        \multirow{8}{*}{$SE(2)$} & \multirow{4}{*}{4}  & \multirow{2}{*}{-} & -       & $83.9 \spm 0.3$   \\
                               &                     &                         & \checkmark   & $81.2 \spm 1.8$    \\
                               &                     & \textit{P}                & -            & $\mathbf{85.1 \spm 0.6}$   \\
                               &                     & \textit{VP}                & -            & $\mathbf{85.1 \spm 0.4}$  \\
                               \cmidrule{2-5}
                               & \multirow{4}{*}{8}  & \multirow{2}{*}{-} & -       & $86.8 \spm 0.6$   \\
                               &                     &                         & \checkmark   & $82.4 \spm 0.5$   \\
                               &                     & \textit{P}                & -            & $\mathbf{87.3 \spm 0.4}$   \\
                               &                     & \textit{VP}                & -            & $\mathbf{87.6 \spm 0.2}$  \\
        \bottomrule
    \end{tabular}    
    }
\end{table}

\paragraph{CIFAR10.} We verify that VP G-CNN for rotation also works well in the widely-used image classification benchmark, CIFAR10. CIFAR10 is a collection of natural object images, such as airplanes, dogs, and so on, and it does not exhibit partial equivariance because the class should not change even if we rotate the image. However, the training and test datasets do not contain rotated images; they only pose upright. This leads partial group convolutions to be partially equivariant. As shown in \cref{tab:test-rotation}, Partial G-CNN and VP G-CNN show competitive performance compared to fully equivariant G-CNN (3rd and 7th rows). This explains that partial equivariance is helpful in CIFAR10. Since the equivariance levels across the data do not differ enough, Partial G-CNN (5th and 9th rows) and VP G-CNN (6th and 10th rows) show comparable performance. On the other hand, InstaAug (4th and 8th rows) presents poor performance even when applied in the regular CNN. This is caused by the unstable training of InstaAug.

\begin{table}[t]
 \setlength{\tabcolsep}{2pt}
\caption{Test accuracy on long-tailed ColorMNIST and Flowers102 with $T(2)\times H_m$ equivariant CNNs. \textit{P} and \textit{VP} denote that their architecture includes Partial and VP convolutional layers, respectively. A \checkmark in the InstaAug column means the training is conducted with the augmentation of InstaAug.}
 \label{tab:test-color}
    \centering
 \resizebox{.45\textwidth}{!}{
    \begin{tabular}{cccccc}\toprule
        Group & \#Elems. & Partial & InstaAug  & ColorMNIST & Flowers102 \\
        \midrule
        \multirow{2}{*}{$T(2) $}                          & \multirow{2}{*}{1}   & \multirow{2}{*}{-}     & -       & $71.0 \spm 0.2$   & $64.6 \spm 0.3$  \\
                                                          &                      &                        & \checkmark    & $70.5 \spm 0.6$   & $66.1 \spm 1.5$  \\
        \midrule
        \multirow{8}{*}{\shortstack[l]{$T(2)$\\$\times H_m$}}& \multirow{4}{*}{3}  & \multirow{2}{*}{-} & -       & $\mathbf{87.1 \spm 0.1}$   & $68.0 \spm 0.5$  \\
                                                           &                     &                         & \checkmark   & $87.3 \spm 0.9$   & $64.3 \spm 1.1$   \\
                                                           &                     & \textit{P}                & -            & $61.4 \spm 0.8$  & $67.2 \spm 1.5$  \\
                                                           &                     & \textit{VP}                & -            & $\mathbf{85.4 \spm 1.0}$  & $\mathbf{69.4 \spm 0.6}$  \\
                                                           \cmidrule{2-6}
                                                           & \multirow{4}{*}{6}  & \multirow{2}{*}{-} & -       & $88.7 \spm 0.3$  & $65.0\spm 0.7$  \\
                                                           &                     &                         & \checkmark   & $87.9 \spm 0.3$  & $62.2 \spm 0.8$  \\
                                                           &                     & \textit{P}                & -            & $63.5 \spm 0.6$  & $66.8 \spm 0.7$  \\
                                                                                                                      &                     & \textit{VP}                & -            & $\mathbf{88.4 \spm 1.1}$  & $\mathbf{69.3 \spm 0.4}$  \\
        \bottomrule
    \end{tabular}    
}
\end{table}

\paragraph{Model architecture for $T(2)\times H_m$.} The product group $T(2)\times H_m$ includes translations and color-shifts. Color Equivariant Convolutions (CEConv) \citep{lengyel2023color} extend the regular CNN, which has $T(2)$ equivariance, to have $H_m$ equivariance. Similar to CEConv, we build ResNet18 consisting of input-aware partial group convolutions of CEConv. We apply the input-aware partial convolution on the last two blocks out of 7 blocks, and the other blocks are all CEConvs with full equivariance. Since CEConv utilizes discrete group elements (3 in \citet{lengyel2023color}) in the Hue spaces, we set $q(u|f)$ as the straight-through distribution as proposed in Equation \cref{eq:ste}. For Flowers102, we also use CEConv to build the fully equivariant G-CNN, but we construct it as the hybrid network consisting of both CEConv and the regular convolutions as in \citet{lengyel2023color}.

\paragraph{Long-tailed colorMNIST.} Long-tailed colorMNIST is a classification task comprising 30 classes of colored digit images. In this task, digits are presented in three different colors against a gray background, requiring classification based on both digits \{0,1,2,3,4,5,6,7,8,9\} and RGB colors \{red, green, blue\}. The distribution of samples per class follows a power law, leading to a significant class imbalance with some classes having substantially more samples than others. Since colorMNIST implies full equivariance for color-shift, as shown in the second last column of \cref{tab:test-color}, CEConv (3rd and 7th rows) shows powerful performance compared to the other baselines, including Partial G-CNN (5th and 9th rows) and InstaAug (4th and 8th rows). Partial G-CNN suffers from unstable training in the discrete groups, thereby it shows poor performance. VP G-CNN (6th and 10th rows), however, provides competitive results, especially when the group elements are 6, which proves the robustness of our model even in the fully equivariant task.

\paragraph{Flowers102.} This dataset consists of 102 different categories of flowers, with each category containing between 40 and 258 images. Each image is labeled with the corresponding category of flower it depicts. As explained in \cref{fig:partial2}, color-shifts of some flowers cause confusion to classifiers, and the range of color-shifts that avoids confusion is non-trivial for each dataset. As seen in the last column of~\cref{tab:test-color}, since G-CNN (3rd and 7th rows) is constructed as a hybrid network, it still works well in such partially equivariant data. On the other hand, Partial G-CNN (5th and 9th rows) still performs poorly due to their instability. Finally, VP G-CNN (6th and 10th rows) significantly outperforms the baselines because of the input-aware partial equivariance and the improved $q(u|f)$ distribution design in discrete groups.

\begin{figure}[t]
     \centering
     \begin{subfigure}[b]{0.23\textwidth}
         \centering
         \includegraphics[width=\textwidth]{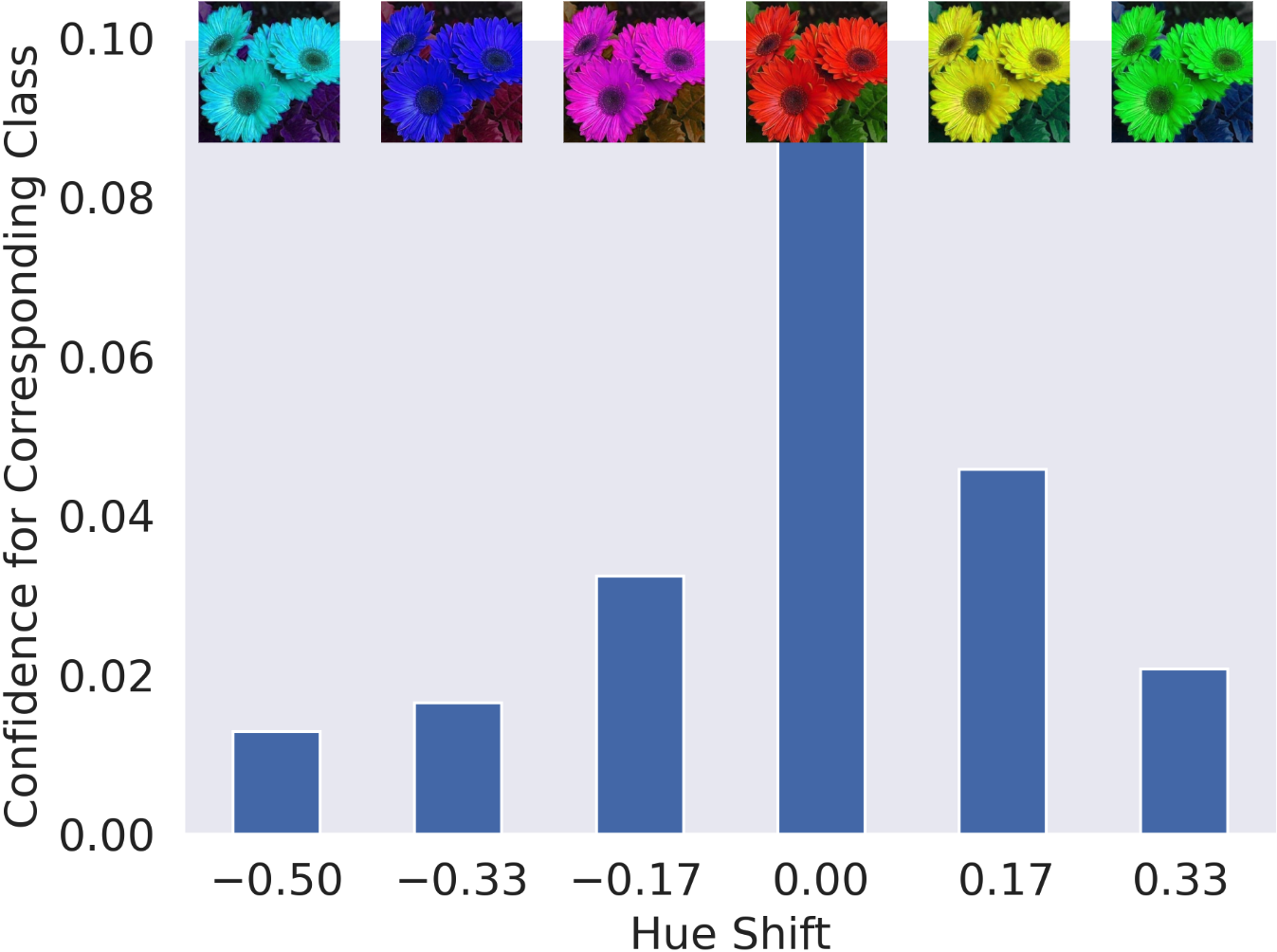}
         \caption{Barberton Daisy}
     \end{subfigure}
     \hfill
     \begin{subfigure}[b]{0.23\textwidth}
         \centering
         \includegraphics[width=\textwidth]{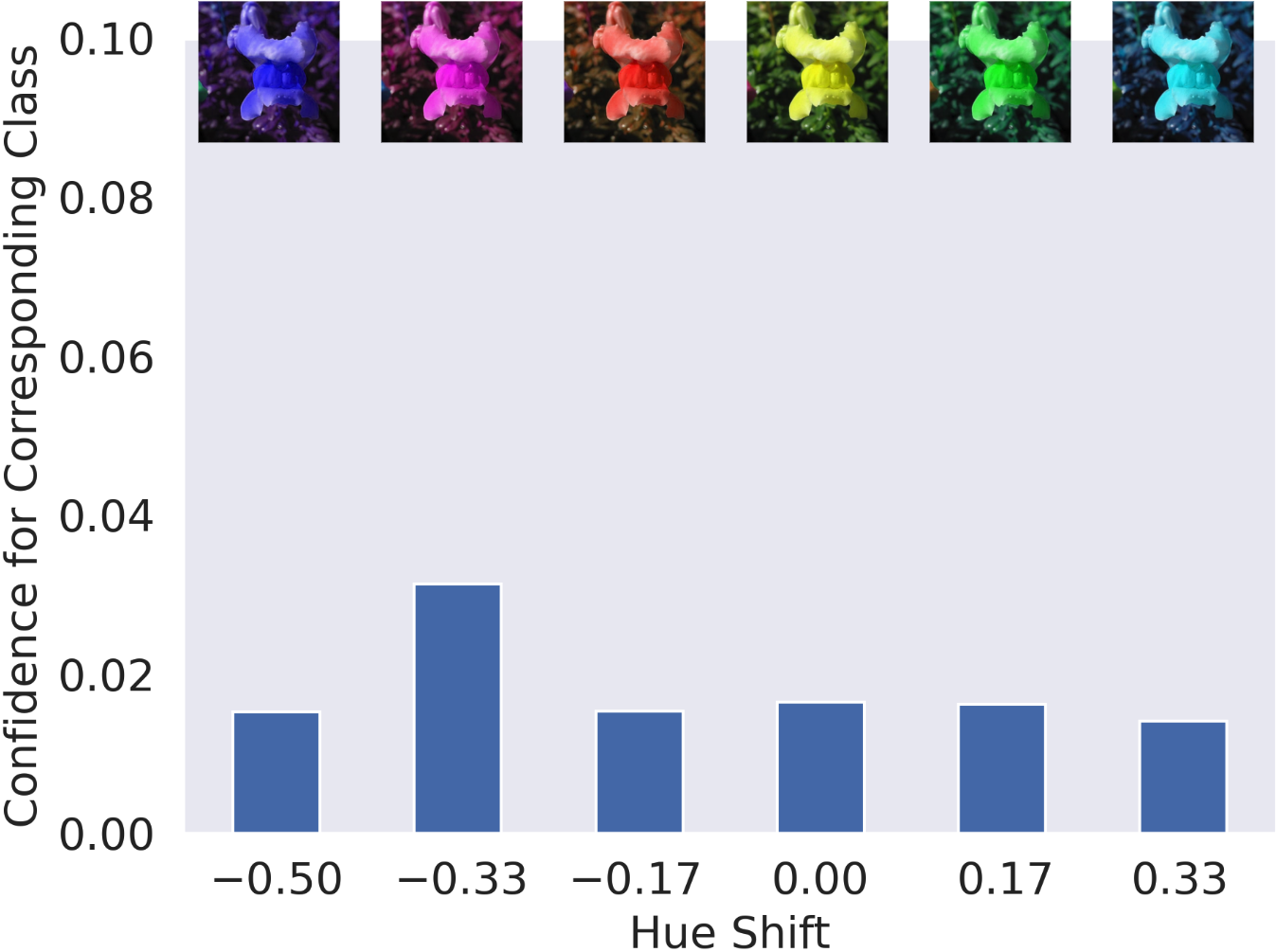}
         \caption{Snapdragon}
     \end{subfigure}
     \vspace{-3mm}
        \caption{The x-axis represents the magnitude of the shift in the Hue space of the input, while the y-axis represents the model's confidence for the corresponding class. The image at zero hue- shift represents the original image. (a) Barberton Daisy exhibits partial equivariance because a shift of -0.17 or 0.17 overlaps with other flowers, while (b) Snapdragon demonstrates full equivariance owing to its distinctive appearance.}
        \label{fig:flowers}
\end{figure}
\paragraph{Learned invariance.}
To verify that our model learns the partial equivariance correctly, we analyze the confidence distribution across the magnitude of the color-shift. \cref{fig:flowers} exhibits the model's confidence for each color-shifted image, and the range of the color-shift is described in $[-0.5,0.5]$. A -0.5 or 0.5 shift produces a complementary color of an image. For Barberton Daisy, the model shows equivariance almost only at 0 shift, while it shows full equivariance for Snapdragon. As explained in \cref{fig:partial2}, if we shift that flower with a -0.17 magnitude, which converts it to purple, it looks like Osteospermum. Conversely, if we shift it with a +0.17 magnitude, which converts it to yellow, it looks like sunflowers, also included in Flowers102. Hence, Barberton Daisy requires no equivariance with respect to the color-shift. On the other hand, due to its unique appearance, Snapdragon is identifiable even if we change its color. Therefore, Snapdragon requires full equivariance, and VP G-CNN captures it properly. The plots for some other flowers can be checked in \cref{fig:flowers-add} of \cref{sec:app:flowers}.

Unfortunately, assessing learned equivariance over the entire dataset through a few metrics can be challenging. Hence, we plotted the minimum and maximum invariance error of the trained model for each class in Flowers102 in \cref{fig:equiv_error} of \cref{sec:app:flowers}. In these plots, the height of bars represents the minimum (or maximum) invariance error across data in each class. Tall bars indicate flowers that require non-equivariance, while short bars represent flowers that require strong equivariance. In the minimum plot, although Tiger Lily (6th from the left) shows non-equivariance, Moon Orchid (7th) and Snapdragon (11th) exhibit relatively strong equivariance. Note that the maximum plot shows that every class includes at least one instance of non-equivariance, indicating that not every flower in a class requires strong equivariance despite their appearance being relatively unique.

\begin{table}[t]
\caption{Test uncertainty metrics on Flowers102.}
 \label{tab:calibration}
    \centering
 \resizebox{.45\textwidth}{!}{
        \begin{tabular}{cccc}
            \toprule
            Metrics & G-CNN & Partial G-CNN & VP G-CNN (Ours)
            \\
            \midrule
            NLL ($\downarrow$) & 0.0171 & 0.0346 & \textbf{0.0121} \\
            BS ($\downarrow$) & 0.0042 & 0.0086 & \textbf{0.0035} \\
            \bottomrule
        \end{tabular}
    }
\end{table}
\paragraph{Calibration performance.} We further compare the calibration performance to evaluate the effectiveness of Variational inference in Flower102. We compute two uncertainty metrics: negative log-likelihood (NLL) and Brier score (BS). NLL represents the discrepancy between the predicted distribution and the actual distribution of the data, quantifying how well the model's predictions match the observed data. On the other hand, BS measures the average squared difference between predicted probabilities and the actual outcomes; lower scores indicate that the predicted probabilities are closer to the actual outcomes. As shown in \cref{tab:calibration}, utilizing Variational inference, VP G-CNN achieves lower NLL and BS scores compared to other methods. This indicates that variational inference is effective not only with respect to accuracy but also in uncertainty quantification.

\paragraph{Stability of proposed discrete distribution.} We compared two discrete distributions for $p(u|f)$ over training time: the Gumbel-Softmax of Partial G-CNN and the Novel Distribution of VP G-CNN. For the same architecture based on VP CEResNet in the Flowers102 task, we only altered the distribution and compared them. That is, the Gumbel-Softmax distribution is also designed to be input-aware by predicting the parameters from the encoder $r_\phi$ as depicted in \cref{fig:noveldist} of \cref{subsec:app:stability}. In each plot, every point represents the probability of each group element $u_1,u_2,u_3$ sampled from $p(u|f)$, and the x-axis denotes the training epochs. For Gumbel-Softmax, the probabilities of each group element frequently vary even at the end of training, while the novel distribution exhibits converged probability distributions (1/3,1/3,1/3) after 300 epochs with minor variations at 575 epochs.

\paragraph{Computational cost.} Since our method requires an extra encoder $r_\phi$ in a few layers to compute the group distribution, additional computational cost is inevitable. \cref{tab:cost} of \cref{sec:app:cost} is a table comparing the computational cost across different methods, in terms of the number of parameters (\#Params) and FLOPs, with CEResNet set as a reference value of 1. CEResNet consists of 1 linear layer, 4 CE residual blocks, and 1 initial CEConv. In our method (VP CEResNet on Flowers102), we replaced one head-side CE residual block (consisting of 3 CEConvs) and one tail-side CEConv with a VP CE residual block and single VP CEConv, respectively. As observed in \cref{tab:cost}, while the number of parameters slightly increases due to the encoder $r_\phi$ utilizing only 1D convolutions, the additional FLOPs are negligible compared to those of CEResNet and Partial CEResNet.

\paragraph{Additional comparisons.} Furthermore, we conduct one additional experiment on CIFAR100, as depicted in \cref{tab:c100} of \cref{subsec:app:c100}, and independently compare our method with another automatic augmentation baseline, AdaAug \citep{cheung2022adaaug}, on Flowers102 as shown in \cref{tab:adaaug} of \cref{subsec:app:adaaug}. We demonstrate that our method competes effectively with other baselines on CIFAR100 and outperforms AdaAug on Flowers102.
\section{Conclusion}
\label{sec:main:conclusion}

We have introduced a new partially equivariant convolution designed to handle partial equivariance encountered in real-world datasets, particularly for groups of rotations and color-shifts. As observed in datasets like MNIST or Flowers102, partial equivariance must be determined based on the input. Unlike the previous Partial G-CNN methods, our approach, named VP G-CNN, learns the appropriate equivariance level for the given input by designing the output group element's distribution in the convolution to be input-aware. We interpret this distribution from a Variational inference perspective as an approximate posterior, enabling us to train the input-dependent distribution less prone to overfitting. Additionally, we have improved the training process, which was unstable in Partial G-CNN, by redesigning the distribution of group elements in the discrete group. We validated our approach on one toy dataset and three real-world datasets, including a fully Equivariant dataset, and confirmed that it captures appropriate partial equivariance for the input while outperforming baseline methods in terms of color equivariance. As an extension of this work, we anticipate that our approach could serve as a guide for constructing a group equivariant network architecture capable of automatically determining the necessary equivariance to a subgroup from a given group, such as the general linear group.


\section*{Acknowledgements}
This research was partly supported by Institute for Information \& communications Technology Promotion(IITP) grant funded by the Korea government(MSIT)(No.RS-2019-II190075, Artificial Intelligence Graduate School Program(KAIST); No.2022-0-00184, Development and Study of AI Technologies to Inexpensively Conform to Evolving Policy on Ethics; No.2022-0-00713, Meta-learning Applicable to Real-world Problems), and the National Research Foundation of Korea(NRF) grants funded by the Korea government(MSIT)(No. 2022R1A5A7083908; No. RS-2023-00279680).

\section*{Impact Statement}
This paper does not include any ethical issues and bad societal consequences. This paper presents a new partial group convolution for mainly image classifications regarding mathematical group symmetry present in data, which does not cause ethical or social issues.


\bibliography{references}
\bibliographystyle{icml2024}

\newpage
\appendix
\onecolumn
\section{Proofs}
\label{sec:app:proofs}

\subsection{Proof of \cref{prop:partial-dist}}
Using change of variables, we expand the convolution integral when the group action $\calL_g$ acts on feature $f$.
\label{proof:partial-dist}
\[
    (k*\calL_g f)(u) &=\int_{v\in G}q(u|\calL_gf)k(v^{-1}u)f(g^{-1}v)\dee\mu_G(v)\\
    &=\int_{v'\in G}q(u|\calL_gf)k({v'}^{-1}g^{-1}u)f(v')\dee\mu_G(v').
\]
On the other hand, the convolution when $\calL_g$ acts on the output of the convolution is
\[
    \calL_g(k* f)(u) &=\int_{v\in G}q(g^{-1}u|f)k(v^{-1}g^{-1}u)f(v)\dee\mu_G(v).
\]
The equivariance error is represented by the difference between the group action on the input and on the output. Then, we bound the $l_2$-norm of the equivariance error using the Cauchy-Schwarz inequality:
\[
    \norm{(k*\calL_g f)(u) - \calL_g (k*f)(u)}_2^2 &= \norm{\int_{v\in G}\Big[q(u|\calL_gf)-q(g^{-1}u|f)\Big]k(v^{-1}g^{-1}u)f(v)\dee\mu_G(v)}_2^2\\
    &\leq\int_{v\in G}\norm{q(u|\calL_gf)-q(g^{-1}u|f)}_2^2\dee\mu_G(v) \int_{v\in G}\norm{k(v^{-1}g^{-1}u)f(v)}_2^2\dee\mu_G(v)\label{eq:equiv-error-bound}\\
    &\leq\begin{cases}
        0,\quad f\in C,\\
        \epsilon^2\cdot\int_{G}\norm{k(v^{-1}g^{-1}u)f(v)}_2^2\dee\mu_G(v),\quad f\in \calF\setminus C
    \end{cases}\label{eq:equiv-error-bound2}
\]
According to the partial equivariance of $q(u|f)$ as in \cref{prop:partial-dist}, the error becomes zero when $f\in C$. Conversely, when $f\in \calF\setminus C$, since the kernel $k$ and input $f$ are bounded, the equivariance error is bounded by a certain value $\epsilon'$. The proof for the lifting convolutions is the same because the integral in \cref{eq:equiv-error-bound2} is still bounded even when integrated over $E$ instead of $G$

\subsection{Proof of \cref{prop:one-layer}}
\label{proof:one-layer}
By mathematical induction, it is enough to show that a G-CNN with two convolution layers, one fully equivariant and another partially equivariant, is itself partially equivariant. For \emph{fully} $G$-equivariant $(k_2*f)(t)$, \emph{partially} $G$-equivariant $(k_1*f)(u)$, $G$-equivariant \& $L$-Lipschitz continuous activation functions $\sigma$, and the bounded kernel $k_2$, the following equivariance error is bounded as:
\[
& \norm{(k_2*\sigma(k_1*\calL_gf))(t)-\calL_g(k_2*\sigma(k_1*f))(t)}^2_2 \\ =& \norm{\big(k_2*\sigma\big[\calL_g(k_1*f)-(k_1*\calL_gf)\big]\big)(t)}^2_2\\
=& \norm{\int_{v\in G} k_2(v^{-1}t)\sigma\big[\calL_g(k_1*f)-(k_1*\calL_gf)\big](v)\dee\mu_G(v)}^2_2\\
\leq& \int_{v\in G} \norm{k_2(v^{-1}t)}^2_2\dee\mu_G(v)\int_{v\in G}\norm{\sigma\big(\calL_g(k_1*f)-(k_1*\calL_gf)\big)(v)}^2_2\dee\mu_G(v)\\
\leq& L^2\int_{v\in G} \norm{k_2(v^{-1}t)}^2_2\dee\mu_G(v)\int_{v\in G}\norm{\big(\calL_g(k_1*f)-(k_1*\calL_gf)\big)(v)}^2_2\dee\mu_G(v).
\]
Thus, the equivariance error is determined by the equivariance error of the partially equivariant convolution $k_1*f$:
\[\norm{\big(\calL_g(k_1*f)-(k_1*\calL_gf)\big)(v)}^2_2,\]
which means if the equivariance error of the partially equivariant network is zero, the equivariance error of the whole network is also zero.

On one hand, for \emph{partially} $G$-equivariant $(k_2*f)(t)$ and \emph{fully} $G$-equivariant $(k_1*f)(u)$ with the associability of the convolution,
\[
\norm{(k_2*\sigma(k_1*\calL_gf))(t)-\calL_g(k_2*\sigma(k_1*f))(t)} &= 
\norm{\Big(k_2*\calL_g\big(\sigma(k_1*f)\big)\Big)(t)-\calL_g(k_2*\sigma(k_1*f))(t)} \\
&=\norm{(k_2*\calL_gf')(t)-\calL_g(k_2*f')(t)},
\] where $f'=\sigma(k_1*f)$, which corresponds to the equivariance error of $k_2*f'$. The proof is still valid when $k_1$ or $k_2$ are the lifting convolution because it is also $G$-equivariant and it has bounded kernels.
\section{Hyperparameters}
\label{sec:app:hyperparameters}

We list hyperparameters used for training VP G-CNN and the baselines in every dataset in~\cref{tab:hyper-all}.
\begin{table}
\begin{subtable}[t]{0.48\textwidth}
    \centering
\newcommand{\metricrule}{\cmidrule(lr){2-3} \cmidrule(lr){4-8}}
\newcommand{\modelrule}{\cmidrule(lr){1-8}}
    \renewcommand{\arraystretch}{1.18}
    \caption{
        The hyperparameter settings used to learn VP G-CNN.
    }
    \vspace{0.5mm}
    \resizebox{0.99\textwidth}{!}{
        \begin{tabular}{lrrrr}
            \toprule
            Dataset & MNIST67-180 & CIFAR10 & ColoredMNIST & Flowers102
            \\
            \midrule
            Batch Size & 64 & 64 & 64 & 64 \\
            Epochs & 300 & 300 & 1500 & 400 \\
            Optimizer & AdamW & AdamW & Adam & Adam \\
            Learning Rate & 0.001 & 0.001 & 0.001 & 0.0002 \\
            Optimizer ($r_\phi$) & SGD & SGD & Adam & AdamW \\
            Learning Rate ($r_\phi$) & 0.001 & 0.001 & 0.0001 & 0.0001 \\
            Weight Decay & 0.001 & 0.001 & 0.00001 & 0.00001 \\
            Architecture Base & ResNet18 & ResNet18 & 7-layer CNN & 7-layer CNN \\
            \# of Conv. Layers & 3 & 5 & 7 & 7 \\
            Normalization & BatchNorm & BatchNorm & BatchNorm & BatchNorm \\
            Kernel Network & SIREN & SIREN & - & - 
            \hspace{3mm}
            \\
            \bottomrule
        \end{tabular}
    }
    \label{tab:hyper}
\end{subtable}
\hfill
\begin{subtable}[t]{0.48\textwidth}
    \centering
\newcommand{\metricrule}{\cmidrule(lr){2-3} \cmidrule(lr){4-8}}
\newcommand{\modelrule}{\cmidrule(lr){1-8}}
    \renewcommand{\arraystretch}{1.18}
    \caption{
        Hyperparameters for baseline models
    }
    \vspace{0.5mm}
    \resizebox{0.9\textwidth}{!}{
        \begin{tabular}{lrrr}
            \toprule
            Dataset & ColoredMNIST & Flowers102 & CIFAR10
            \\
            \midrule
            Batch Size & 256& 64 & 64  \\
            Epochs & 1500 & 300 & 300\\
            Optimizer & Adam & Adam & Adam \\
            Learning Rate & 0.001  & 0.001 & 0.001  \\
            Weight Decay &  0.00001 & 0 & 0.0001  \\
            Architecture Base & 7-layer CNN & ResNet18 & ResNet18 \\
            Kernel Network & - & - & SIREN \\
            Optimizer\textsuperscript{*}  & Adam & Adam & AdamW \\
            Learning Rate\textsuperscript{*} & 0.0001 & 0.0001 & 0.0001  \\
            Optimizer\textsuperscript{**}  & Adam & Adam & Adam \\
            Learning Rate\textsuperscript{**} & 0.0001 & 0.0001 & 0.001  \\
            Entropy regularization \textsuperscript{**} & 0.0001 & 0.0001 & 0.0001  \\
            \bottomrule
            \small * PG-CNN specific parameters \\
            \small ** InstaAug specific parameters
        \end{tabular}
    }
    \label{tab:hyper-base}
\end{subtable}

\caption{Hyperparameter settings.}
\label{tab:hyper-all}
\end{table}
\section{Architecture of Encoder $e_\phi$}
\label{sec:app:encoder}

\begin{figure}[h!]
    \centering
    \includegraphics[width=0.5\textwidth]{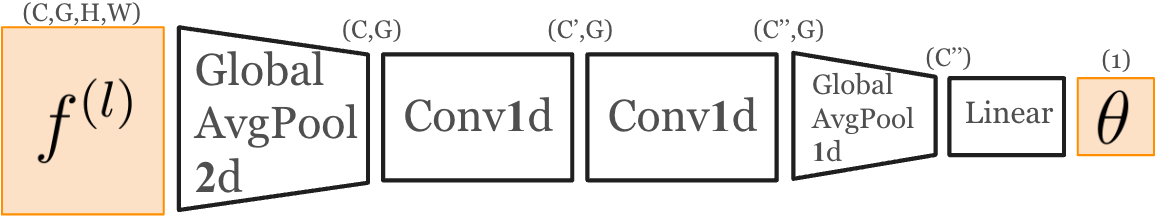}
    \caption{Encoder network $e_\phi(f)$ architecture.}
    \label{fig:encoder}
\end{figure}

We use light-weighted encoder $e_\phi$, consisting of two global average pooling layers, two one-dimensional convolution, and one linear layer. The encoder is illustated in~\cref{fig:encoder}. The parentheses above the network indicate dimensions of input tensors in each layer. $C,C',C'',C''$ denotes the number of channels, $G$ is the number of group elements, $H$ is height of the features, and $W$ is width of the features.

\section{Additional Plots in Flowers102}
\label{sec:app:flowers}

\begin{figure}[h!]
     \centering
     \begin{subfigure}[b]{0.23\textwidth}
         \centering
         \includegraphics[width=\textwidth]{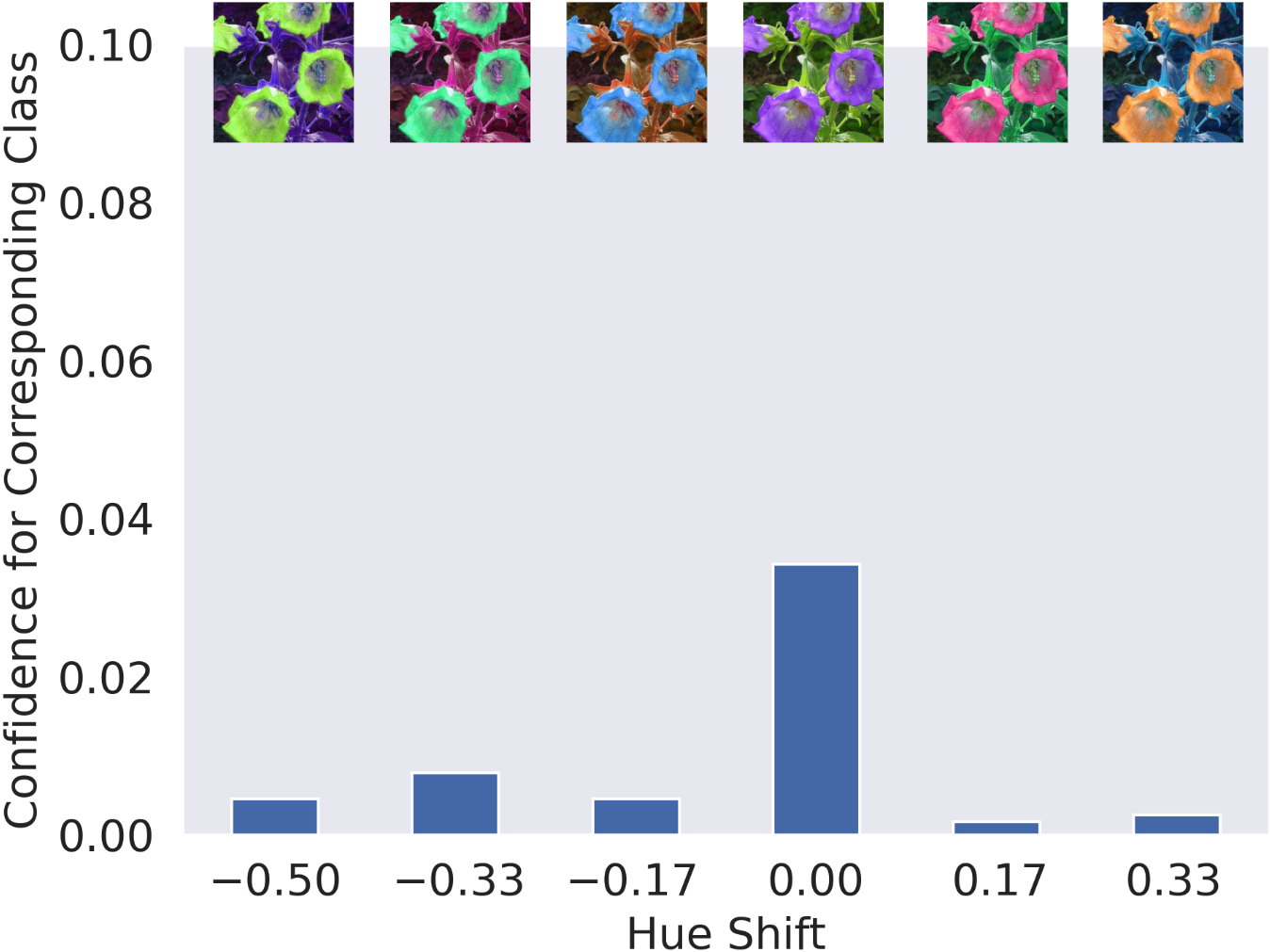}
         \caption{Canterbury Bells}
     \end{subfigure}
     \hfill
     \begin{subfigure}[b]{0.23\textwidth}
         \centering
         \includegraphics[width=\textwidth]{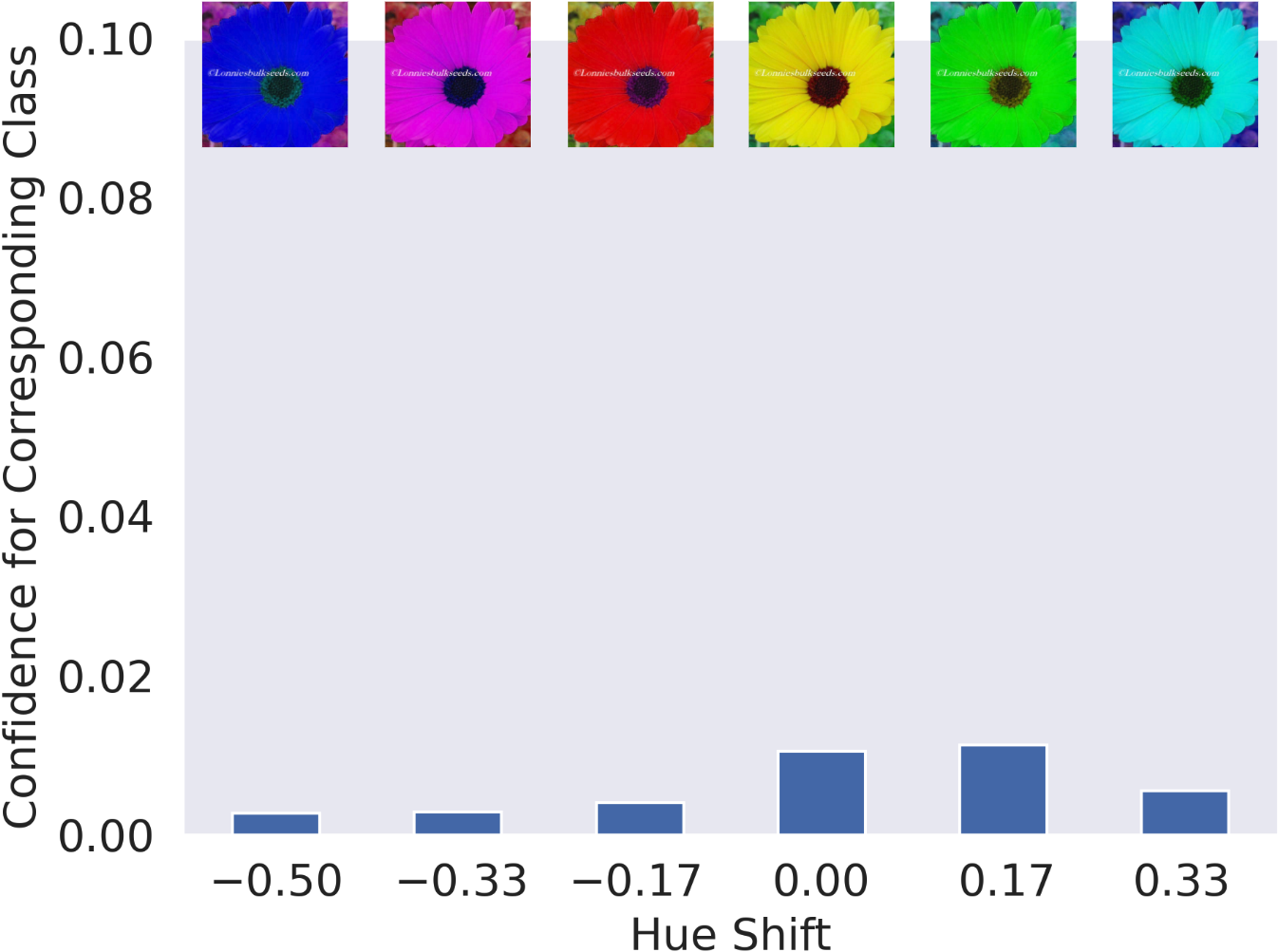}
         \caption{English Marigold}
     \end{subfigure}
     \hfill
     \begin{subfigure}[b]{0.23\textwidth}
         \centering
         \includegraphics[width=\textwidth]{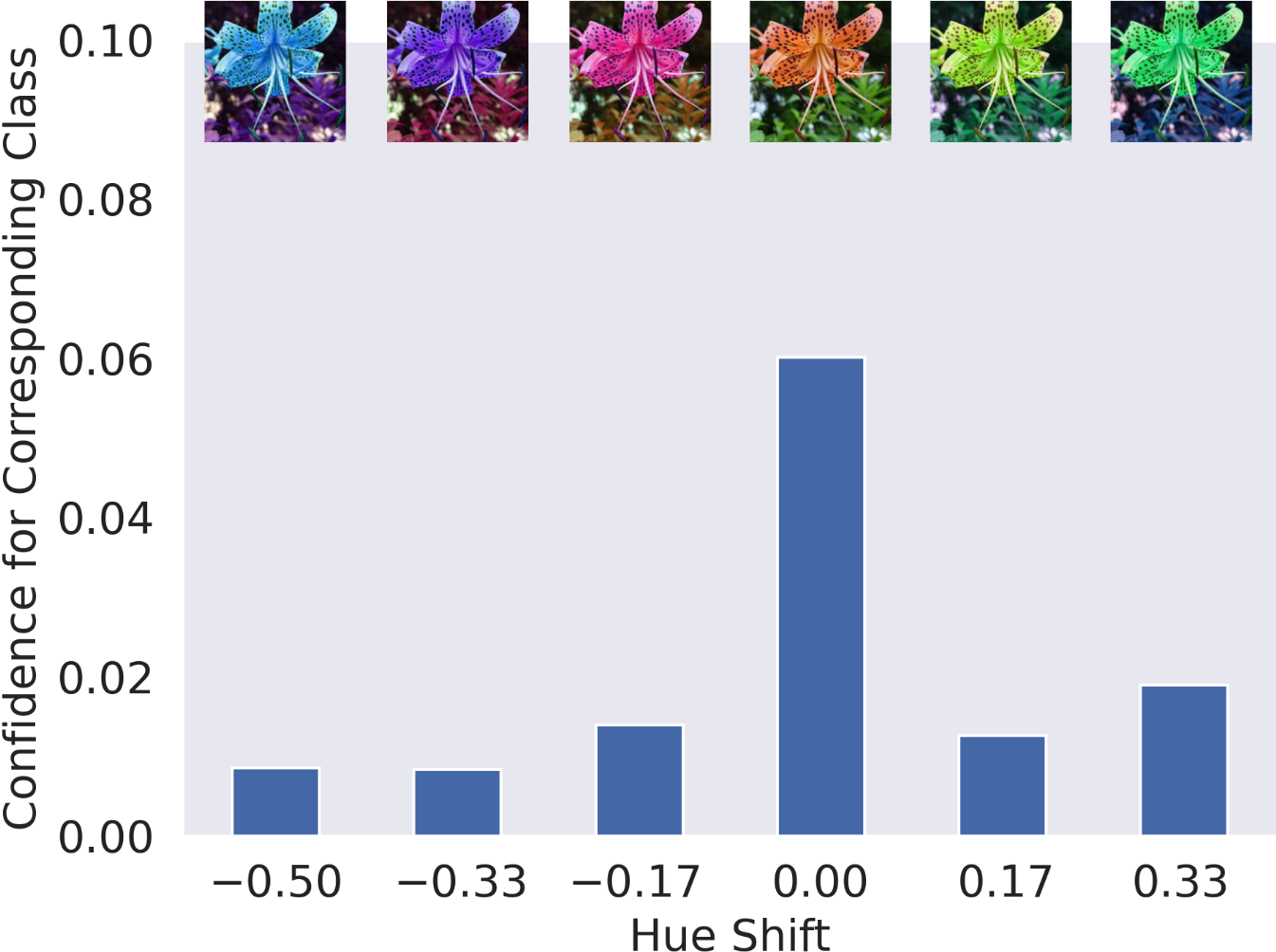}
         \caption{Tiger Lily}
     \end{subfigure}
     \begin{subfigure}[b]{0.23\textwidth}
         \centering
         \includegraphics[width=\textwidth]{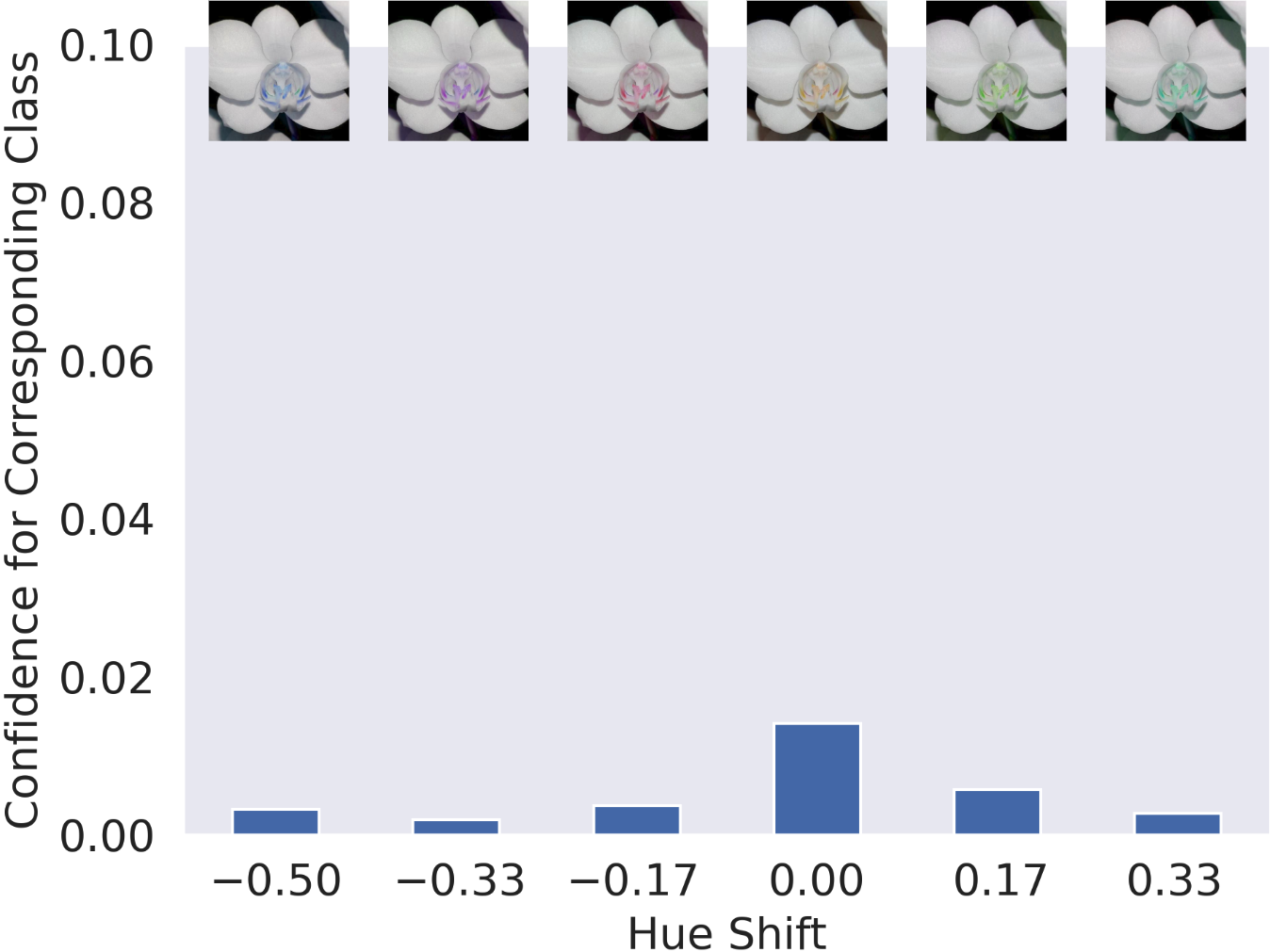}
         \caption{Moon Orchid}
     \end{subfigure}
     \vspace{-3mm}
        \caption{Confidence across color-shifts of input image in Flowers102.}
        \label{fig:flowers-add}
\end{figure}

\begin{figure}[h!]
    \centering
    \begin{subfigure}[b]{\textwidth}
        \centering
        \includegraphics[width=\textwidth]{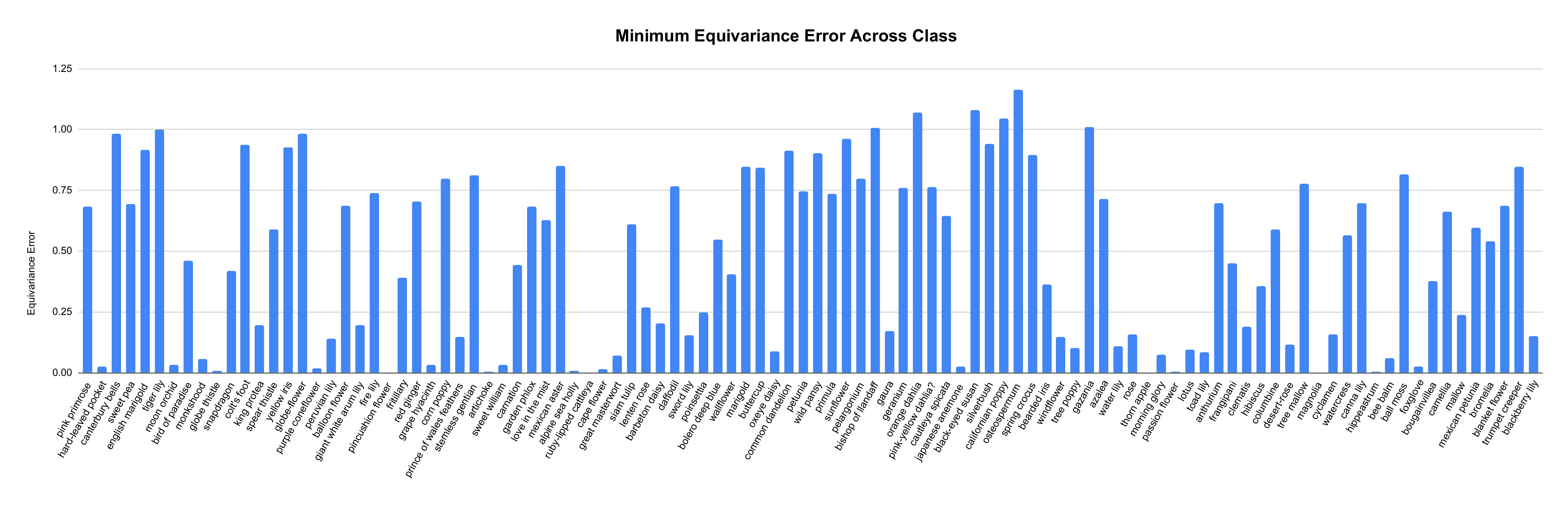}
        \caption{Minimum}
    \end{subfigure}\hfill
    \begin{subfigure}[b]{\textwidth}
        \centering
        \includegraphics[width=\textwidth]{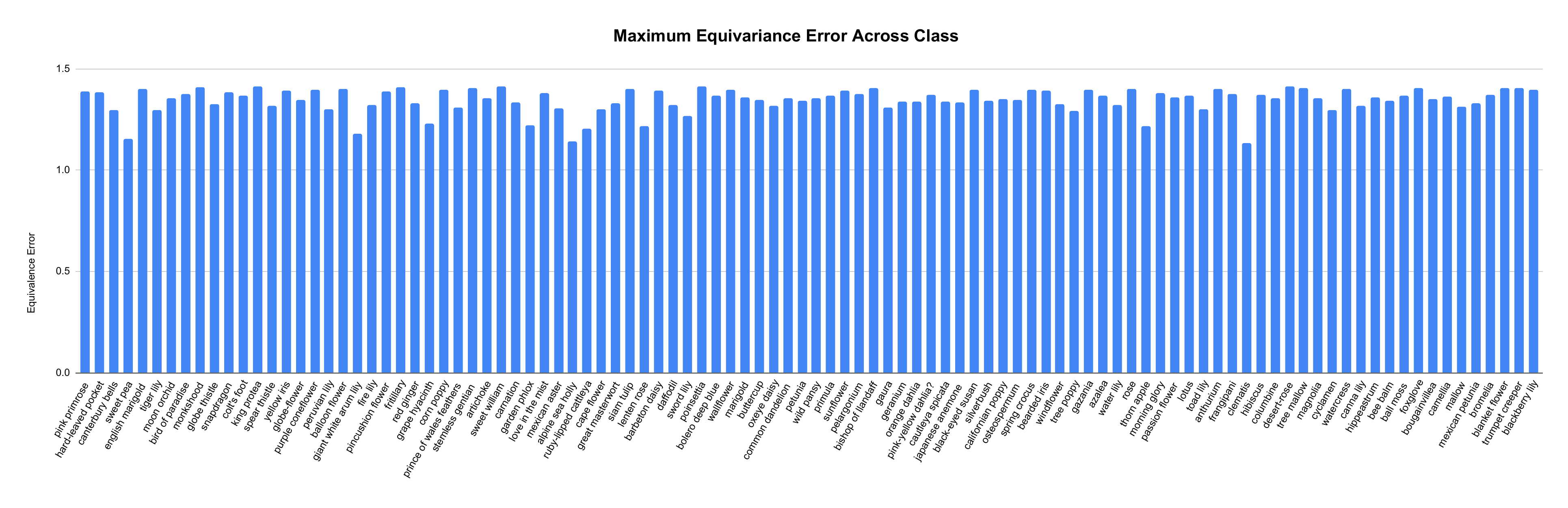}
        \caption{Maximum}
    \end{subfigure}
    \caption{Mimimum and Maximum equivariance errors across classes in Flowers102. Tall bars indicate flowers that require non-equivariance, while short bars represent flowers that require strong equivariance.}
    \label{fig:equiv_error}
\end{figure}
\section{Computational Cost}
\label{sec:app:cost}

Since our method requires an extra encoder $r_\phi$ in a few layers to compute the group distribution, additional computational cost is inevitable. Below is a table comparing the computational cost across different methods, in terms of the number of parameters (\#Params) and FLOPs, with CEResNet set as a reference value of 1. CEResNet consists of 1 linear layer, 4 CE residual blocks, and 1 initial CEConv. In our method (VP CEResNet on Flowers102), we replaced one head-side CE residual block (consisting of 3 CEConvs) and one tail-side CEConv with a VP CE residual block and single VP CEConv, respectively.

\begin{table}[h!]
\caption{Computational cost comparison of models used in Flowers102. * denotes the model reported in \cref{tab:test-color}. $\dag$ indicates a model whose layers are all VP.}
 \label{tab:cost}
    \centering
 \resizebox{\textwidth}{!}{
        \begin{tabular}{ccccccc}
            \toprule
            Metrics & CEResNet & ResNet w/ InstaAug & Partial CEResNet & Partial CEResNet w/ InstaAug & VP CEResNet\textsuperscript{*} & VP CEResNet$^\dag$
            \\
            \midrule
            \#Params ($\downarrow$) & $\times$1.0000 & $\times$1.9628 & $\times$1.0000 & $\times$1.9814 & $\times$1.2416 & $\times$1.3211 \\
            FLOPs ($\downarrow$) & $\times$1.0000 & $\times$0.3161 & $\times$1.0000 & $\times$1.1581 & $\times$1.0006 & $\times$1.0007 \\
            \bottomrule
        \end{tabular}
    }
\end{table}

As observed in \cref{tab:cost}, while the number of parameters slightly increases due to the encoder $r_\phi$ utilizing only 1D convolutions, the additional FLOPs are negligible compared to those of CEResNet and Partial CEResNet.
\section{Additional Experiments}
\label{sec:app:additional}

\subsection{Evaluation in CIFAR-100}
\label{subsec:app:c100}

We conducted experiments using the partially $SE(2)$-equivariant ResNet on the CIFAR100 dataset. As you can see in \cref{tab:c100}, our model still performs competitively on CIFAR100. However, the improvement is not surprising compared to Partial G-CNN because CIFAR100 is not necessarily trained to be aware of the level of rotation invariance for each data point.
\begin{table}[h!]
\caption{Test accuracy in CIFAR100.}
 \label{tab:c100}
    \centering
 \resizebox{0.7\textwidth}{!}{
        \begin{tabular}{cccc}
            \toprule
              & $SE(2)$-ResNet & Partial $SE(2)$-ResNet & VP $SE(2)$-ResNet (ours)
            \\
            \midrule
            Test Accuracy (\%;$\uparrow$) & 52.40 & 57.02 & \textbf{57.67} \\
            \bottomrule
        \end{tabular}
    }
\end{table}
Furthermore, our performance remains constrained at around 50\% as we utilized the same model architecture as in the experiments conducted in the Partial G-CNN paper, resulting in weaker performance compared to contemporary models.

\subsection{Comparison with AdaAug}
\label{subsec:app:adaaug}

We compared our method with AdaAug \citep{cheung2022adaaug} on Flowers102. AdaAug is a method for learning adaptive data augmentation policies in a class-dependent and potentially instance-dependent manner. It trains the policy network that determines the augmentations via the validation loss evaluated by the augmented validation data and the classifier, while training the classifier with the augmented training data. Here is the test accuracy comparison with different baselines used with AdaAug that generates adaptive color-shift augmentations as shown in \cref{tab:adaaug}.

\begin{table}[h!]
\caption{Test accuracy comparison with AdaAug in Flowers102.}
 \label{tab:adaaug}
    \centering
 \resizebox{\textwidth}{!}{
        \begin{tabular}{ccccc}
            \toprule
              & ResNet w/ AdaAug & CEResNet w/ AdaAug & Partial CEResNet w/ AdaAug & VP CEResNet (ours)
            \\
            \midrule
            Test Accuracy (\%;$\uparrow$) & 64.70 & 63.51 & 68.45 & \textbf{69.40} \\
            \bottomrule
        \end{tabular}
    }
\end{table}

Although AdaAug is fairly a good method, our method still outperforms it due to its architectural inductive bias. Furthermore, our method does not demand the validation dataset.

\subsection{Stability of Proposed Discrete Distribution}
\label{subsec:app:stability}

\begin{figure}[h]
    \centering
    \includegraphics[width=\textwidth]{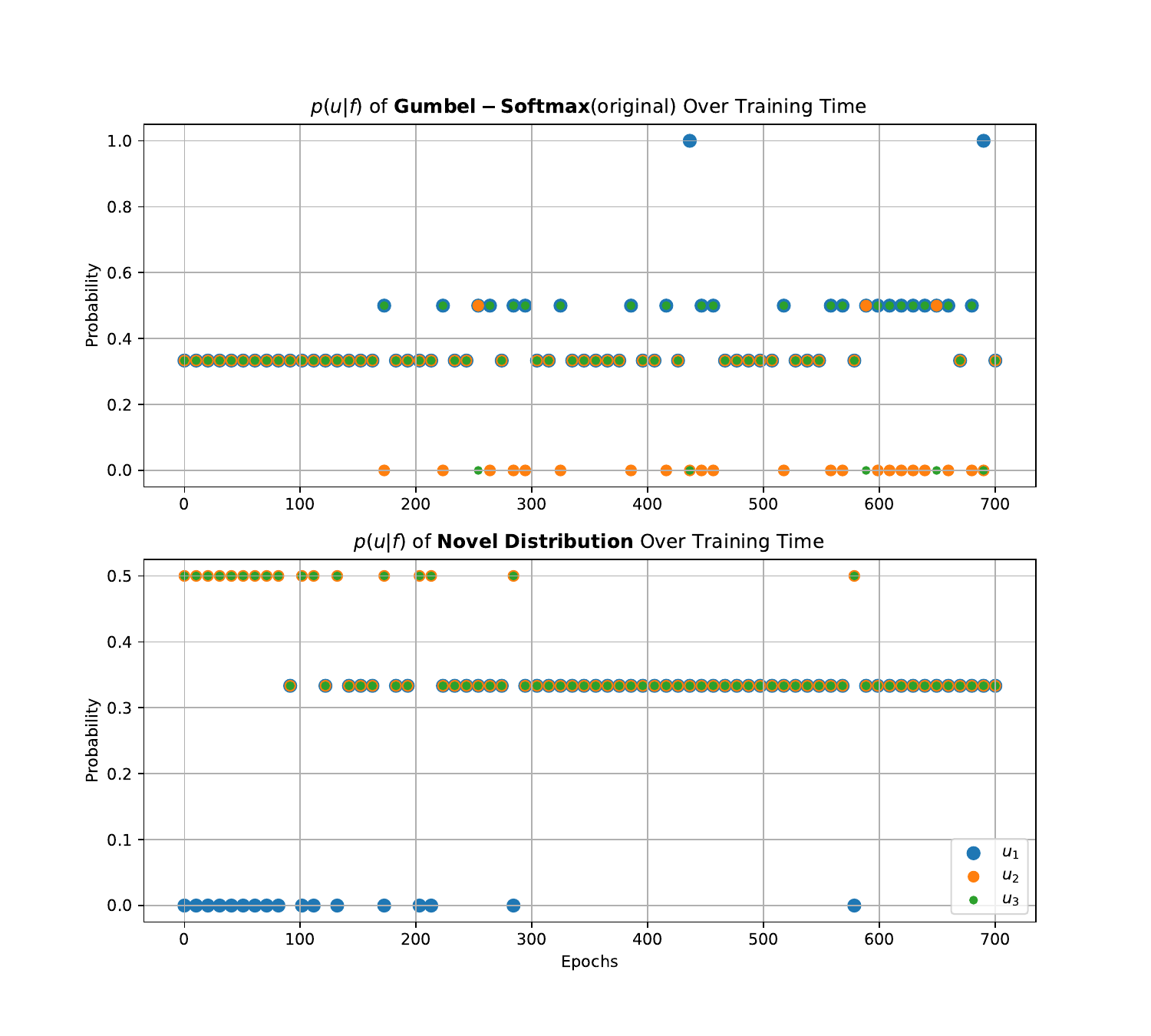}
    \vspace{-9mm}
    \caption{Gumbel-Softmax (top) vs. Novel Distribution (bottom)}
    \label{fig:noveldist}
\end{figure}

We compared two discrete distributions for $p(u|f)$ over training time: the Gumbel-Softmax of Partial G-CNN and the Novel Distribution of VP G-CNN. For the same architecture based on VP CEResNet in the Flowers102 task, we only altered the distribution and compared them. That is, the Gumbel-Softmax distribution is also designed to be input-aware by predicting the parameters from the encoder $r_\phi$ as depicted in \cref{fig:noveldist}.

In each plot, every point represents the probability of each group element $u_1,u_2,u_3$ sampled from $p(u|f)$, and the x-axis denotes the training epochs. For Gumbel-Softmax, the probabilities of each group element frequently vary even at the end of training, while the novel distribution exhibits converged probability distributions (1/3,1/3,1/3) after 300 epochs with minor variations at 575 epochs.

\end{document}